\documentclass[graybox]{svmult}

\usepackage{layouts}
\usepackage{type1cm}        %
\usepackage{makeidx}         %
\usepackage{graphicx}        %
\usepackage{multicol}        %
\usepackage[bottom]{footmisc}%

\newcommand{\ie}{\emph{i.e.,~}}
\newcommand{\figref}[1]{Fig.~\ref{#1}}
\usepackage{subcaption}
\definecolor{groundcolor}{HTML}{E05215}
\definecolor{shrubcolor}{HTML}{6C7A66}
\definecolor{treecolor}{HTML}{44AF1D}
\definecolor{stemcolor}{HTML}{AD9A0B}
\definecolor{canopycolor}{HTML}{132A14}
\definecolor{ForestGreen}{RGB}{34,139,34}

\definecolor{DeciduousColor}{HTML}{ff00c5}
\definecolor{ConiferousColor}{HTML}{005ce6}
\definecolor{MixedColor}{HTML}{aaff00}

\newcommand{\DeciduousSquare}{{\textcolor{DeciduousColor}{$\blacksquare$}}}
\newcommand{\ConiferousSquare}{{\textcolor{ConiferousColor}{$\blacksquare$}}}
\newcommand{\MixedSquare}{{\textcolor{MixedColor}{$\blacksquare$}}}

\usepackage{booktabs} 
\usepackage{multirow}

\usepackage{todonotes}

\usepackage{newtxtext}       %
\usepackage{newtxmath}       %
\usepackage[backend=bibtex,maxcitenames=1]{biblatex}
\usepackage{siunitx}
\usepackage{tabularx}
\usepackage{array}
\usepackage{microtype}

\newcolumntype{Y}{>{\centering\arraybackslash}X}
\newcolumntype{Z}{>{\centering\arraybackslash}m{2cm}}  %
\newcolumntype{L}{>{\raggedright\arraybackslash}m{2cm}}  %

\newcolumntype{P}[1]{>{\raggedright\let\newline\\\arraybackslash\hspace{0pt}}m{#1}}
\newcolumntype{C}[1]{>{\centering\let\newline\\\arraybackslash\hspace{0pt}}m{#1}}

\usepackage{hyperref}

\makeindex             %

\begin{document}
\title*{DigiForest: Digital Analytics and Robotics for Sustainable Forestry}
\author{Marco Camurri, Enrico Tomelleri, Mat{\'i}as Mattamala, Sebastián Barbas Laina, Martin Jacquet, Jens Behley, Sunni Kanta Prasad Kushwaha, Fang Nan, Nived Chebrolu, Leonard Frei{\ss}muth, Marvin Chayton Harms, Meher V.R. Malladi, Fan Yang, Jonas Frey, Cesar Cadena, Marco Hutter, Janine Schweier, Kostas Alexis, Cyrill Stachniss, Maurice Fallon, Stefan Leutenegger}
\institute{Marco Camurri \at Dipartimento di Ingegneria Industriale, University of Trento, via Sommarive 9, 38123 Trento (TN), Italy \email{marco.camurri@unitn.it} \at Faculty of Engineering, Free University of Bozen-Bolzano, via Bruno Buozzi 1, 39100 Bozen-Bolzano (BZ), Italy
\and Enrico Tomelleri \at Faculty of Agricultural, Environmental and Food Sciences, Free University of Bozen-Bolzano, piazza Università 5,  39100 Bozen-Bolzano (BZ), Italy \email{enrico.tomelleri@unibz.it} \at Competence Centre for Mountain Innovation Ecosystems, via Bruno Buozzi 1, 39100 Bozen-Bolzano (BZ), Italy
\and Mat{\'i}as Mattamala \at School of Informatics, University of Edinburgh, 10 Crichton St, EH8 9AB, United Kingdom \email{matias.mattamala@ed.ac.uk}
\and Sebastián Barbas Laina, Leonard Frei{\ss}muth \at Technical University of Munich, School of Computation, Information and Technology
Boltzmannstrasse 3, 85748 Garching, Germany \email{{sebastian.barbas, l.freissmuth}@tum.de}
\and Martin Jacquet, Marvin Chayton Harms, Kostas Alexis \at Department of Engineering Cybernetics, Norwegian University of Science and Technology, Elektro D/B2, Gløshaugen, Norway. \email{{martin.jacquet, marvin.c.harms, konstantinos.alexis}@ntnu.no}
\and Jens Behley, Meher V.R. Malladi, Cyrill Stachniss \at Center for Robotics, University of Bonn, Nussallee 15, 53115 Bonn, Germany \email{{jens.behley, meher.malladi, cyrill.stachniss}@igg.uni-bonn.de}
\and Sunni Kanta Prasad Kushwaha, Janine Schweier  \at Swiss Federal Research Institute WSL, Zürcherstrasse 111, 8903 Birmensdorf, Switzerland \email{{sunni.kushwaha, janine.schweier}@wsl.ch}
\and Fang Nan, Fan Yang, Jonas Frey, Cesar Cadena, Marco Hutter \at Robotic Systems Lab, ETH Z{\"u}rich, Leonhardstrasse 21, 8092 Z{\"u}rich, Switzerland \email{{fannan, fanyang1, jonfrey, cesarc, mahutter}@ethz.ch}
\and Nived Chebrolu, Maurice Fallon \at Oxford Robotics Institute, Department of Engineering Science, University of Oxford, 23 Banbury Rd, Oxford,  OX2 6NN,  United Kingdom \email{{nived, mfallon}@robots.ox.ac.uk}
\and Stefan Leutenegger, Leonard Frei{\ss}muth \at Mobile Robotics Lab, ETH Zürich, Department of Mechanical and Process Engineering, Leonhardstrasse 21,
8092 Zürich, Switzerland \email{{lestefan, lfreissmuth}@ethz.ch}
} %
\maketitle

\abstract*{Covering one third of Earth's land surface, forests are vital to global biodiversity, climate regulation, and human well-being. In Europe, forests and woodlands reach approximately 40\% of land area, and the European forestry sector is central to achieving the EU’s climate neutrality and biodiversity goals, which emphasizes sustainable forest management, increased use of long-lived wood products, and resilient forest ecosystems. To meet these goals, current practices require further innovation to meet rising environmental demands. This chapter introduces DigiForest, a novel, large-scale precision forestry approach leveraging digital technologies and autonomous robotics. DigiForest is structured around four main components: (1) autonomous, heterogeneous mobile robots (aerial, legged, and marsupial) for tree-level data collection; (2) automated extraction of tree traits to build forest inventories; (3) a Decision Support System (DSS) for forecasting forest growth and supporting decision-making; and (4) low-impact selective logging using purpose-built autonomous harvesters. These technologies have been extensively validated in real-world conditions in several locations, including forests in Finland, the UK, and Switzerland.}

\abstract{Covering one third of Earth's land surface, forests are vital to global biodiversity, climate regulation, and human well-being. In Europe, forests and woodlands reach approximately 40\% of land area, and the forestry sector is central to achieving the EU’s climate neutrality and biodiversity goals; these emphasize sustainable forest management, increased use of long-lived wood products, and resilient forest ecosystems. To meet these goals and properly address their inherent challenges, current practices require further innovation. This chapter introduces DigiForest, a novel, large-scale precision forestry approach leveraging digital technologies and autonomous robotics. DigiForest is structured around four main components: (1) autonomous, heterogeneous mobile robots (aerial, legged, and marsupial) for tree-level data collection; (2) automated extraction of tree traits to build forest inventories; (3) a Decision Support System (DSS) for forecasting forest growth and supporting decision-making; and (4) low-impact selective logging using purpose-built autonomous harvesters. These technologies have been extensively validated in real-world conditions in several locations, including forests in Finland, the UK, and Switzerland.}

\section{Introduction}
\label{sec:intro}

Forests cover nearly one third of the Earth’s land surface and are essential for biodiversity, climate regulation, and several other ecosystemic services. In Europe the share is even higher (close to 40\%) and, as outlined by the EU Forest Strategy for 2030~\cite{EU2021}, forestry has a key role in reaching EU's climate neutrality and biodiversity goals, calling for more efficient and sustainable forest management.

Forest management practices differ from country to country, but generally follow a four-stage cycle. The first is surveying: given a forest plot or stand of interests, foresters record absolute position, Diameter at Breast Height (DBH), height, species, and health status of each tree, as well as additional traits of specific interest. Typically, collection of such quantities is performed manually in the field with portable tools (e.g., calipers for DBH and laser range finders for height) \cite{foresthandbook}. These surveys are generally accurate, but are subject to human operator's ability, slow to carry out, and sparse (only limited areas are covered).
More recently, techniques such as terrestrial laser scanning (TLS) \cite{lidarremotesensing}, mobile laser scannning (MLS) \cite{bauewns2016forests} \cite{slam2007harvester}, and aerial laser scanning (ALS) are used to collect detailed 3D information on tree and stand structure. MLS has been demonstrated to be superior in treat estimation such as DBH, as coverage per unit time outweighs sampling accuracy \cite{bauewns2016forests}.

The second stage involves the aggregation of the data collected to create (or update) a forest inventory.  A forest inventory keeps track of all the relevant traits required for forest planning, such as DBHs, species and health. Inventories are made on-site during the data collection itself, or offline by means of specialized software to extract tree traits from point clouds \cite{treeseg}, also aggregating longitudinal information.

The third stage is planning. Based on the inventory, foresters decide which interventions are more suitable for a stand, e.g., thinning, harvesting, or other measures to promote regeneration. Depending on the application, the aim may be to stimulate growth, reduce competition, increase structural diversity, or secure timber production. Forest planning has traditionally relied on expert knowledge and paper records, but modern planning systems, known as Decision Support Systems (DSS), integrate computational models to estimate growth and production in different scenarios.

The fourth and final stage is carrying out forest interventions, which are determined by the planning step. Current forestry, specially in European countries, is shifting away from timber production via clear cuts in favor of continuous cover forestry. This approach aims to preserve ecological diversity while still targeting the production goals. Such effort requires the coordination of different activities such as thinning, pruning, planting, tending regeneration, apart from protecting against fire, browsing, and pests.

Among these, thinning is a silvicultural intervention in which a portion of trees is periodically removed to reduce competition and promote the growth, stability, and quality of the residual stand. By favoring well-formed trees and reducing density, thinning enhances the value of the timber, increases the resilience to biotic and abiotic stresses, and can improve biodiversity through greater structural heterogeneity. In European managed forests, thinning is typically scheduled at several stages during a rotation and adapted to site conditions and management objectives. Because the whole cycle is time and resource intensive, management usually runs on a five to ten year cycle.

With climate change and rising frequency and intensity of disturbance \cite{patacca2023gcb}\cite{pomerlau2020jfr}, there is growing pressure to shorten these cycles and make forest management more flexible. While modern remote sensing techniques provide large-scale aerial forest data to improve decision-making, over-canopy measurements lack the fidelity to improve current growth and production models. Therefore, there is an urging need for precision forestry solutions, which prioritize collecting fine-grained, localized under-canopy data while employing technological solutions that minimize the environmental impacts of this industry.

\begin{figure}
    \centering
    \includegraphics{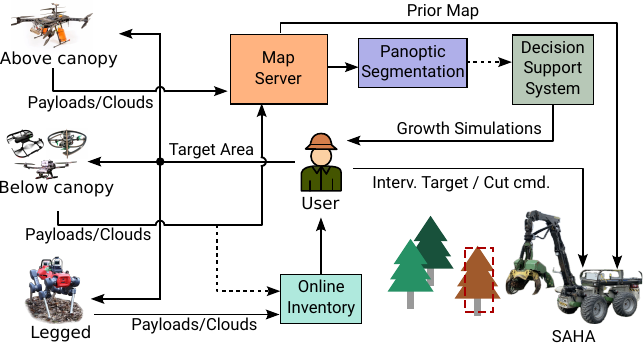}
    \caption{Overview of the forest management approach proposed by DigiForest and data flow (solid lines indicate demonstration in the field, dashed lines indicate future integrations). The user initiate the data collection by providing a target area to survey. The maps from the ground units produce standardized payload maps for online tree inventory or multi-session mapping communicating them to a map server via a standardized interface. The maps are used to perform panoptic segmentation to produce tree traits, including its components such as crown and shrub, as well as to provide a prior map for the autonomous harvester. A decision support system uses the tree traits to produce growth simulation and predict the effect of an intervention. Finally, the operator decides the intervention to make and authorize the cut when the autonomous harvester has reached the site of intervention.
    }
    \label{fig:digiforest-overview}
\end{figure}

\begin{figure}
    \centering
    \includegraphics[width=0.5\linewidth]{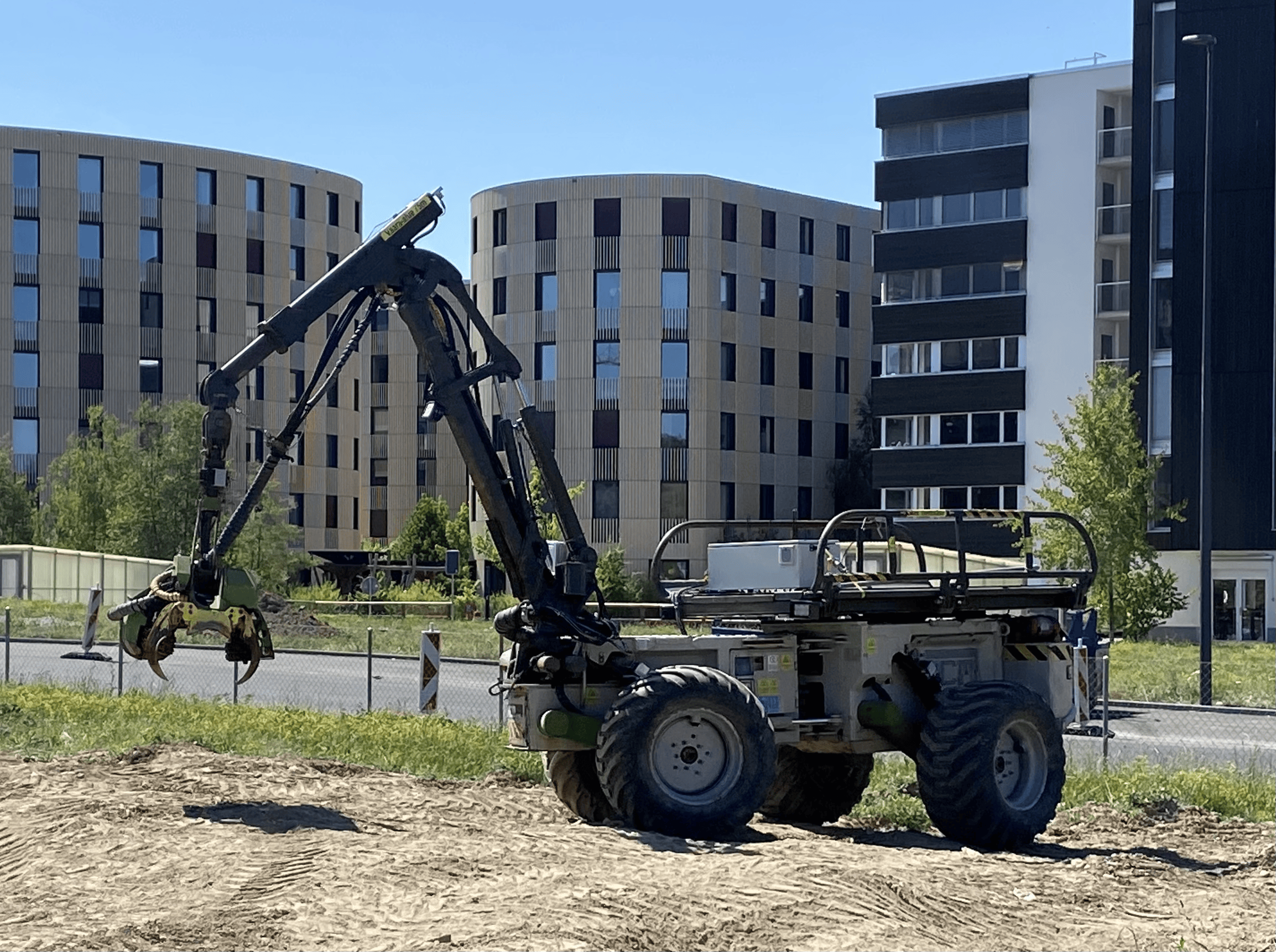}
    \caption{The SAHA autonomous harvester.}
    \label{fig:SAHA}
\end{figure}

In this work, we present an innovative approach towards large-scale automated precision forestry based on digital analysis and robotic technologies developed in the context of the DigiForest EU Horizon project\footnote{https://digiforest.eu/}.
An overview of DigiForest is shown in \figref{fig:digiforest-overview}. Inspired by the four stages of current forestry practices, the approach is centered around four main components:
\begin{enumerate}
    \item A detailed, tree-level data collection pipeline triggered by a human operator and executed by autonomous heterogeneous mobile robots. In addition to above-canopy ALS data acquisition, we focus on automated ground-level scanning by means of legged robots and small scale drones. While wheeled platforms have been used for this task in the past \cite{pierzchala2018compag} \cite{pomerlau2020jfr}, legged robots are now a mature technology promising superior mobility and reduced soil impact than wheeled and tracked vehicles. This technology is complemented by autonomous undercanopy aerial drones, which can effectively complement legged platforms in ground data acquisition as they can reach dense undergrowth areas but have more limited battery life. The platforms exploit their sensing capabilities to reconstruct the forest via SLAM online or by communicating the data in a standardized format \cite{payloadapi} to a map server running multi-session, multi-platform SLAM.
    \item An automated tree trait extraction and online forest inventory for immediate report to the user in the field, or more detailed LiDAR processing capable of identifying trees as well as its constitutive components via panoptic segmentation. In contrast to most MLS or TLS based systems, the inventory is available online during the mission. The inventory system was demonstrated on the legged robots which were able to cover 1 ha in 20-30 min with DBH estimation accuracy of 2-3 cm.
    \item A Decision Support System (DSS) performing growth forecast from forest inventory data, to support operator decision making.
    \item Automated selective logging/cutting with small footprint, purpose-built, hydraulic harvesting machine. In conventional thinning, harvesting is a two-step cycle: (i) felling and processing trees along extraction trails, and (ii) forwarding the assortments to the roadside with a separate machine. Although effective, this approach relies on heavy harvesters and forwarders (15–20~t), which concentrate loads on forest soils. Repeated passes along trails can lead to rutting and soil compaction with long-term impacts on soil erosion and stand productivity. There is therefore a clear need for lighter, more versatile solutions that combine cutting and short-distance forwarding in a single machine, minimizing soil disturbance while operating in confined stands.
To meet these needs, we developed the SAHA robot (\figref{fig:SAHA}) a 4.5-ton harvester optimized for selective thinning tasks, based on the RCM Harveri~\cite{jelavic2022harveri}
\end{enumerate}
These components have been designed to operate independently from each other, such that each one could be used interchangeably with the forest management practice currently in use.

The remainder of the chapter is structured as follows. Section \ref{sec:autonomous-forest-navigation} describes the autonomy and mapping  of the quadruped robot, as well as the autonomy stack for undercanopy flight, which includes a novel submapping based with trajectory adaptation upon loop closure, as well as a Neural Control Barrier Function (CBF) and MPC for reactive replanning in case of drift. Section \ref{sec:automated-forest-inventory} describes the online inventory system that was fully integrated with the quadruped platform, as well as the panoptic segmentation technique to segment trees and their subcomponents. Section \ref{sec:dss} introduces an initial case study for the Decision Support System (DSS) which includes a growth simulator to support management decisions. Section \ref{sec:harvester} describes the semi-autonomous harvesters, including automation, navigation and grasping/cutting capabilities. Section \ref{sec:experimental} shows the experimental assessment of the systems presented in prior sections. Finally, Section \ref{sec:conclusion} closes the chapter with final remarks and considerations for future works.

\section{Autonomous Forest Navigation and Mapping}
\label{sec:autonomous-forest-navigation}
In this section, we describe the autonomous navigation and mapping system of our heterogeneous team of robots (legged, aerial, marsupial).
Each robot category has specific complementary characteristics: legged platforms can cross a variety of obstacles and have long endurance, but cannot traverse through tall and dense undergrowth. Conversely, aerial platforms can rapidly navigate under-canopy but have limited payload and battery life.
As each platform has its characteristic capabilities and requirements, we divide the remainder of the section by platform type. A summary of the platforms is shown in Table \ref{tab:digiforest-platforms}.

\begin{table}[ht]
\centering
\begin{tabularx}{\textwidth}{LYYZY}
\toprule
\textbf{Name} & \textbf{Type} & \textbf{Dimension [cm]} & \textbf{Weight [kg]} & \textbf{Battery Life [min]} \\
\midrule
Avartek Boxer & UAV & 105 $\times$ 105 $\times$ 82 & ~17 (30 max) & 120 \\
\hline
RMF & UAV / Marsupial & 38 $\times$ 38 $\times$ 24 & 1.45 & 8 \\
\hline
MRL & UAV & 38 $\times$ 38 $\times$ 24 & 1.7 & 15 \\
\hline
BLK2FLY & UAV & 53 $\times$ 60 $\times$ 19 & 2.6 & 10 \\
\hline
ANYmal & Legged / Marsupial & 93 $\times$ 53 $\times$ 80 & 50 & 90--120 \\
\hline
SAHA & Harvester & 422 $\times$ 220 $\times$ 281 & 4500 & 960+ \\
\bottomrule
\end{tabularx}
\caption{Overview of DigiForest robotic platforms.}
\label{tab:digiforest-platforms}
\end{table}

\subsection{Legged Robot Navigation}
\label{sec:forest-inventory-legged}

\begin{figure}[tb]
    \centering
      \includegraphics[width=0.8\linewidth]{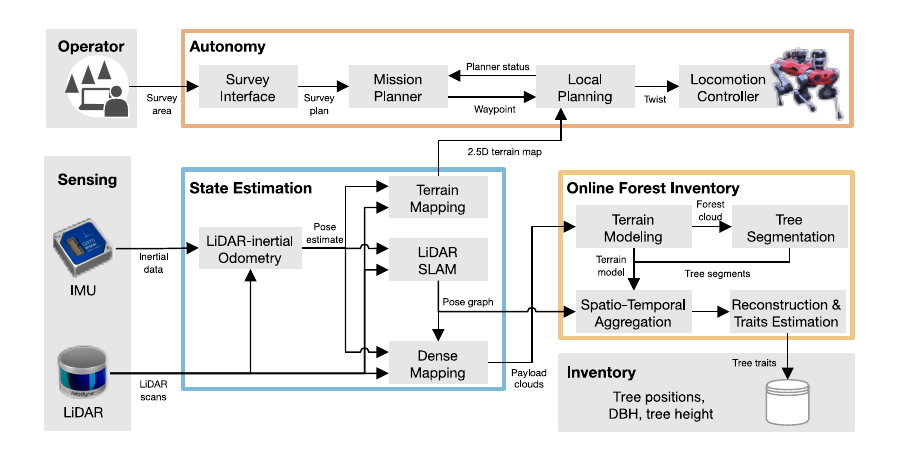}
      \caption{System overview. The \textbf{Autonomy system} executes the mission specified by a survey reference provided by a human operator. \textbf{State estimation} provides a consistent scene representation and dense clouds used for estimating tree traits. \textbf{Forest inventory} segments the point cloud data and estimates tree traits from point cloud data. The main output of the system is a forest inventory database with the main attributes of the surveyed forest.
      }
      \label{fig:anymal-system-overview}
\end{figure}

For ground data acquisition, we developed and deployed an autonomy and forest inventory software stack on the ANYmal D quadruped platform. An overview of the architecture is shown in Fig.~\ref{fig:anymal-system-overview}: our solution incorporates a state-of-the-art LiDAR-based odometry and SLAM system \cite{oh2024iros}, a mission planning and navigation system, and an online forest inventory pipeline, which we describe in Section \ref{subsec:forest-inventory}. Further details and results from previous field trials in Finland and the UK are described in a prior work~\cite{mattamala2025tfr}.

The ANYmal robot’s navigation system combines a reactive local planner with a global path-planning strategy. The local planner handles real-time obstacle avoidance and adapts to terrain changes. It outputs $SE(2)$ twist commands, which the robot's lower-level control executes. Meanwhile, the global planner ensures long-term goal tracking and mission coverage. The planner leverages both geometric features (e.g., slope, elevation, obstacles) and semantic information (e.g., grass, mud) to make informed decisions in forest environments. This integration enables the robot to navigate safely and efficiently without becoming stuck (behind trees or obstacles) or deviating from its path.

\subsubsection{Legged State Estimation}
\label{sec:state-estimation}
For state estimation (blue box in Fig.~\ref{fig:anymal-system-overview}), we used a LiDAR-inertial factor graph-based odometry solution, VILENS~\cite{Wisth2023}, which provides high-frequency pose and velocity estimates ($\sim$\SI{200}{\hertz}) in a fixed inertial frame. The system performs sensor fusion of relative pose displacements from LiDAR using the ICP algorithm~\cite{Besl1992}, aided by preintegrated IMU measurements \cite{Forster2017}. We calibrated the cameras, as well as Camera-IMU extrinsics using the Kalibr tool \cite{furgale2013iros}. In addition, we developed a method for calibration LiDAR and camera using a checkerboard \cite{fu2023iros}.

The factor graph-based LiDAR SLAM system from~\cite{Ramezani2020a} is used to obtain a consistent estimate of the robot pose in a fixed global frame. Our SLAM system incrementally builds a pose graph, detects loop closure candidates within a radius of \SIrange{10}{15}{\meter} from the robot, and optimizes the pose graph in an incremental fashion using the iSAM2 algorithm~\cite{Kaess2012}.

We implemented a local sub-mapping module to accumulate all the incoming LiDAR scans within a fixed-distance window, generating dense maps we refer to as \emph{data payloads} \cite{payloadapi}. These data payloads are gravity-aligned and published at a lower frequency ($\sim$\SI{0.1}{\hertz}), accumulating clouds for about every \SI{20}{\meter} traveled.

The system also perform local terrain mapping and use its output as the main representation for local planning~\cite{Miki2022b}. We used a 2.5D grid-based representation with a resolution of \SI{4}{\centi\meter}.

\subsubsection{Autonomy System}
\begin{figure*}[t]
\centering
  \includegraphics[width=0.75\linewidth]{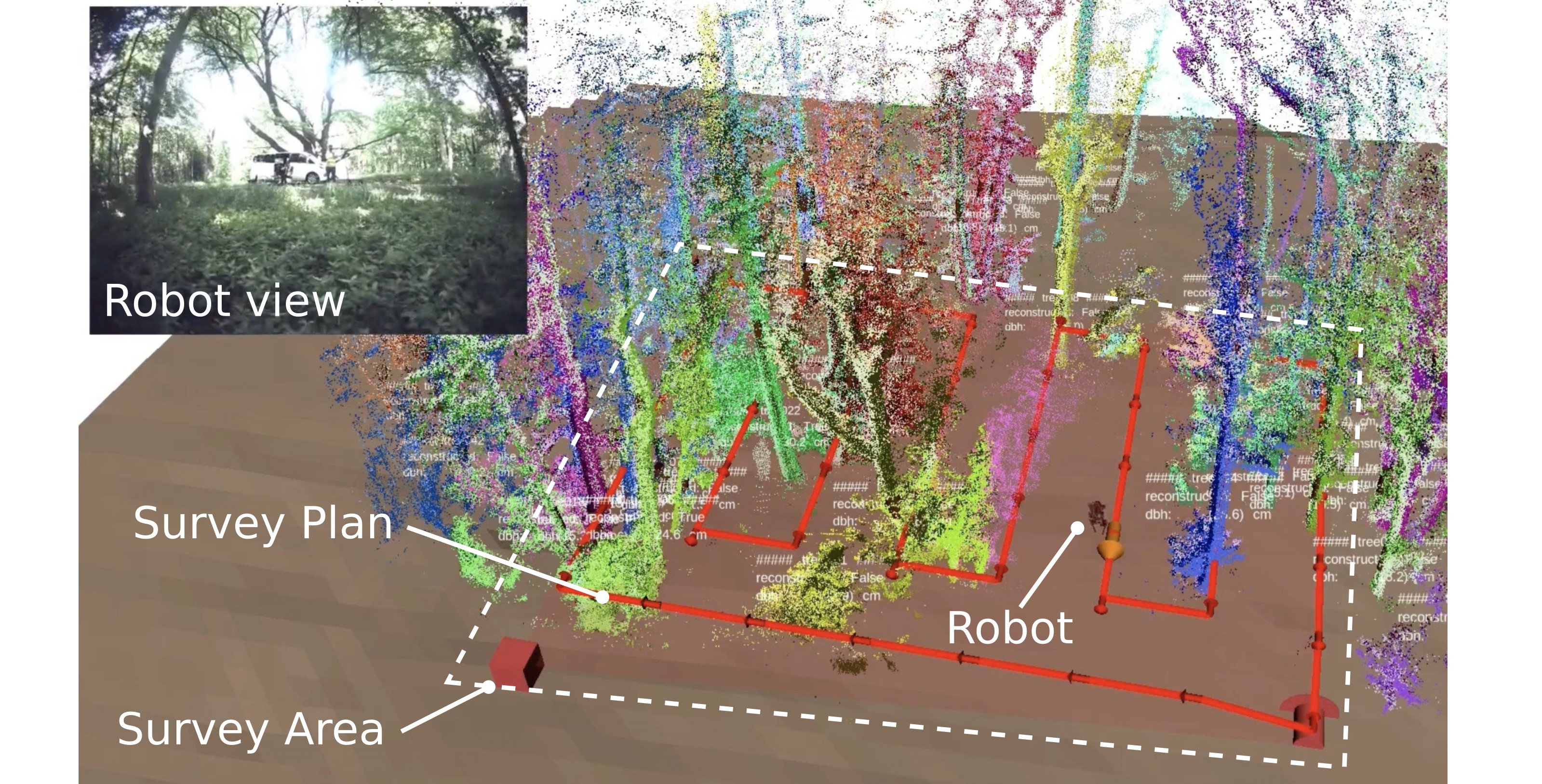}
  \caption{Survey Interface: GUI for the operator, implemented on RViz using interactive markers. The operator defines the area to be surveyed, while the survey plan is automatically proposed by the Mission Planner. The segmented trees are shown in different colors.}
  \label{fig:alf-mission-planning}
\end{figure*}

The input to the Autonomy System (orange box in \figref{fig:anymal-system-overview}) comes from the human operator, who defines the robot deployment plan from a graphical user interface (GUI) based on RViz using interactive markers (Fig.~\ref{fig:alf-mission-planning}). Given the robot's initial pose, the GUI allows the operator to define a survey area and initialize a \emph{survey plan} given by a \emph{lawn mower pattern}, as done for TLS data collection protocols~\cite{Duncanson2020}.

Once the survey plan from the operator is received, the mission planner manages its execution. The main task of the mission planner is scheduling the waypoints to be followed by the local planner, and executing any high-level action from the operator, such as safety stops. The local planner generates velocity commands for the low-level locomotion controller to reach a specified waypoint. We used a reactive local planner~\cite{Mattamala2022}, which exploits the local terrain map to determine the action the robot executes.

The local planner outputs a 3 DoF linear and angular velocity command for a low-level RL-based locomotion controller. In our experiments, we used both the perceptive controller presented by Miki et al.~\cite{Miki2022a}, as well as the blind, learning-based controller from ANYbotics.

\subsection{Under-canopy Safe Navigation and Mapping}
\label{sec:undercanopy-flight}
Umanned Aerial Vehicles (UAV) for data acquisition above-canopy are becoming more commonly used in the forestry sector, as they provide a more flexible alternative to fixed-wing aircraft-mounted sensors for Aerial Laser Scanning~(ALS) surveying. We employ a GNSS waypoint navigation plan to collect ALS data from an Avartek Boxer Hybrid UAV, to capture structural details that cannot be obtained from terrestrial scanning. The LiDAR collected from above-canopy is then fused with the data from under-canopy using the factor-graph based markerless approach described in \cite{casseau2024iros}.

In contrast to above-canopy ALS acquisition, autonomous under-canopy navigation is more challenging due to the presence of irregular obstacles, such as trunks and low branches, and the unreliability of GNSS data.
To study the problem of safe navigation under-canopy in a general and versatile way, we tested our algorithms on three UAV systems: the MRL drone, which only relies on visual-inertial sensing, a collision-tolerant Resilient Micro Flyer (RMF) \cite{depetris2020msc}, and a commercial LiDAR-based drone, the Leica BLK2FLY.

\begin{figure}
    \centering
    \includegraphics[width=\linewidth]{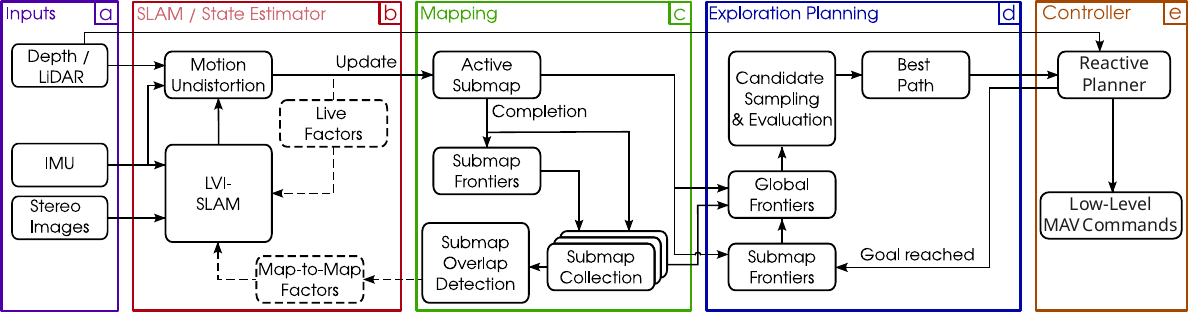}
    \caption{Overall system architecture for autonomous exploration and mapping. The system leverages stereo images, IMU data and depth information from either a LiDAR or stereo images which are used for state estimation (b). The mapping module generates locally consistent submaps, that are stored together with the submaps' local frontiers (c). These are later used in Exploration Planning (d) to compute global frontiers. After evaluating the frontiers for the best next drone pose, the path planner plans a trajectory to it, which the Model Predictive Controller (MPC) converts into attitude angles to control the UAV (e). The method can optionally leverage the submaps' information to improve the state estimation accuracy.}
    \label{fig:okvis_system_overview}
\end{figure}

The under-canopy navigation pipeline has four main components, illustrated in Fig.~\ref{fig:okvis_system_overview}: a) a tightly coupled depth-visual-inertial SLAM, b) a large-scale volumetric mapping, c) an autonomous exploration and navigation process that relies on the volumetric maps, and d) a local collision avoidance controller.

\subsubsection{Tightly coupled Depth-Visual-Inertial SLAM with Large-Scale Volumetric Occupancy Mapping}
\label{sec:mav-slam}

The UAV state estimation algorithm (Fig.~\ref{fig:okvis_system_overview}.b) is based on OKVIS2 \cite{leutenegger2022okvis2}, a visual-inertial system which has been extended to integrate depth information to build dense volumetric occupancy submaps for both geometric reconstruction and path planning. The depth information can come from a LiDAR sensor or stereo vision approaches, such as \cite{xu2023unifying}.

The visual keypoints extracted from stereo cameras are tightly coupled with temporal IMU measurements to perform odometry estimates. The state estimator uses the poses to integrate depth information into the volumetric submaps, which are then used for downstream tasks. At the same time, the system can be configured to incorporate map-to-map and depth-to-map residuals into the state estimator, improving the odometry accuracy at the expense of higher computational requirements.

If the state estimator is configured to leverage the submap's geometric information, we modify the factor-graph optimization as from our previous work \cite{bocheLVIOkvis}. In the Stein am Rhein field trials, we tested the proposed system and compared the reconstruction quality obtained by enabling and disabling the submap constraints. During the tests, the depth was obtained from either a LiDAR sensor (tested with the BLK2FLY) or from a Realsense D455 sensor (MRL UAV). All computations from the state estimator ran online.

\subsubsection{Scalable Autonomous Drone Flight in the Forest with Visual-Inertial SLAM and Dense Submaps}

For mapping (Fig.~\ref{fig:okvis_system_overview}.c), the autonomous navigation system uses the same submaps from the estimator described in Section \ref{sec:mav-slam} for safe path planning. These submaps are constructed using the Supereight2 \cite{funk2021multi} multi-resolution occupancy mapping framework, where each voxel stores the log-odds probability of being occupied. From this probability, regions within the submap are classified as free, occupied, or unknown. Unknown regions correspond to areas that are unobserved.

For path planning, we use the Informed-RRT* algorithm \cite{gammell2014informed} and evaluate candidate path segments across all available submaps. A segment is considered valid if it lies entirely within free space in at least one submap. If the segment intersects an occupied or unknown region in a given submap, we check subsequent submaps. To ensure safety, we adopt a conservative approach by treating unknown regions as occupied during planning.

\begin{figure}[t]
	\centering
	\includegraphics[width=.7\columnwidth]{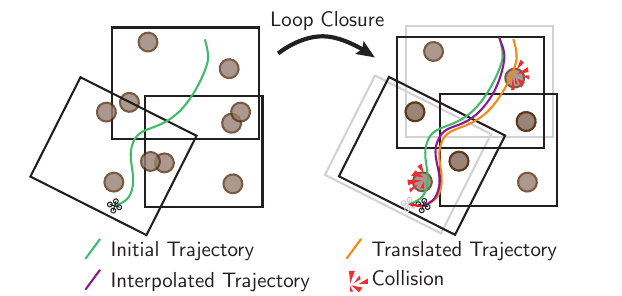}
	\caption{
		Planned trajectory before loop-closure (left) and trajectory adaptation strategies after loop-closure (right), with obstacles as brown disks. The adaptation strategies are: no adaptation (green), rigid transformation (orange) and the proposed trajectory anchoring (purple).	Without adaptation, the UAV flies towards its estimated pose before loop-closure (red arrow), which is unsafe. Rigidly transforming the trajectory does not account for the accumulated drift of each submap along it, leading to unsafe trajectories.
	}
	\label{fig:trajectory_anchoring_simu}
	\vspace{-0.3cm}
\end{figure}

Once a safe path is found, it is converted into a trajectory that accounts for the UAV’s velocity and acceleration constraints. This trajectory is anchored dynamically using updates from the state estimator. Since the estimator includes drift-correction mechanisms like loop closures, the trajectory must deform in real-time to remain consistent with the corrected state. Without this adaptation, discontinuities from loop closures could lead to erratic behavior or crashes, as the UAV might attempt to reach a trajectory point that now corresponds to a different physical location, as shown in \figref{fig:trajectory_anchoring_simu}. To address this issue, we deform the planned path based on changes in the underlying submaps, allowing the trajectory to elastically adapt and remain within known free space.

\subsubsection{Efficient Submap-based Autonomous UAV Exploration using Visual-Inertial SLAM Configurable for LiDARs or Depth Cameras}
\label{sec:submap_mav_exploration}

Since the objective of the autonomous exploration algorithm is to visit previously unseen areas, we exploit the log-odds probability of occupancy to identify candidates to be explored.

We begin by describing how the algorithm operates in a monolithic map before introducing the additional processes required for a submapping-based approach, as proposed in \cite{papatheodorou2024efficient}. The exploration planner (\figref{fig:okvis_system_overview}.d) generates candidate positions near \emph{map frontiers} \ie regions likely to expand the known map if observed. In a volumetric occupancy map, a frontier is defined as the boundary between known free space and unobserved space.

From the UAV’s current position, a path is planned to each candidate location. The estimated time to reach each candidate, $t_j$, is computed assuming the UAV travels at its maximum linear and angular velocities. At each candidate position, a 360$^\circ$ entropy-based raycast is performed to estimate the expected information gain, producing a \emph{gain image}. Based on the sensor’s horizontal field of view, a sliding window optimization is applied to identify the yaw angle that yields the highest gain $g_j$. The utility of each candidate is then calculated as $u_j = g_j / t_j$, effectively prioritizing candidates that maximize information gain per unit time. The candidate with the highest utility is selected as the next exploration goal (\figref{fig:okvis_system_overview}.e) and the process repeats once the goal has been reached.

\begin{figure}[t]
	\centering
	\includegraphics[width=.7\columnwidth]{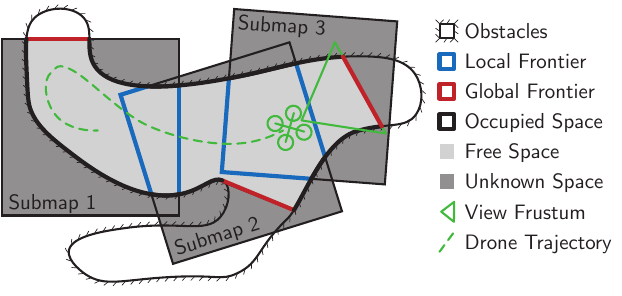}
	\caption{Local-only (blue) and global (red) frontiers in a set of submaps. The local-only submap frontiers correspond to regions mapped in other submaps and are thus not considered to be global frontiers.}
	\label{fig:frontiers}
\end{figure}

In the submapping framework, the utility computation and raycasting steps remain the same but are applied over the entire collection of submaps. The primary difference lies in how global frontiers are defined for sampling candidate positions. Within a single submap, a frontier is any region where free space borders with unknown space or lies along the submap’s boundary. However, in a multi-submap system, a frontier in one submap may no longer be valid in another if that region has already been observed. To handle this, we evaluate frontiers after a submap is completed. Using the known pose transformations $T_{WM_k}$ for each submap, we test whether the local frontiers from one submap still correspond to unknown space in others. If the region is classified as free or occupied in any other submap, it is excluded as a sampling frontier. An illustrative example is shown in \figref{fig:frontiers}.

\subsubsection{Local Replanning for Collision Avoidance}
Volumetric exploration planners provide paths that are nominally collision-free and typically rely on a voxel representation with a coarse resolution, so as to enable real-time computation. However, a coarse resolution makes it difficult to capture thin branches and twigs, posing risks of collisions and potential damage to the environment and/or the robot. Moreover, the volumetric map relies solely on the SLAM software and is therefore highly sensitive to systematic errors, such as position drift and estimation inaccuracies.

Consequently, it is important to introduce additional layers for collision avoidance by designing a complementary safety control layer that does not ``blindly'' track the planned trajectory. We therefore developed a safety layer able to process exteroceptive sensor data and assess whether actions are safe. Sensor input may come from an onboard LiDAR or a high-resolution depth camera, both of which provide detailed geometric information about the immediate surroundings.

Specifically, we use two methods designed to be used in a cascaded manner.
The first method is a Nonlinear Model Predictive Controller (NMPC) exploiting a neural Signed Distance Field~(SDF) as an explicit position constraint. It generates acceleration commands for tracking the reference trajectory.
The second method relies on Control Barrier Functions~(CBFs) to provide a last-resort safety filtering of the acceleration commands without relying on neural networks.

Each method has distinct advantages and trade-offs. The use of neural networks in the NMPC allows to work directly with high-dimensional sensor data. The method ensures a smooth behavior through predictive planning, but is computationally expensive. In contrast, the CBF is a reactive, computationally efficient formal approach with certified safety.

\begin{figure}[t]
    \includegraphics[width=\linewidth]{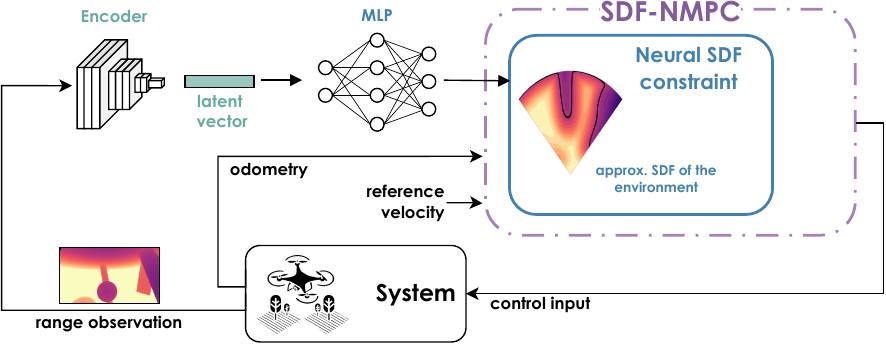}
    \caption{A block-diagram of the SDF-NMPC method.
        The range image is processed via a neural network to encodes the corresponding SDF within the sensor frustum. This SDF is used to parametrize a constraint in the NMPC, ensuring that trajectories satisfy a minimum distance to the obstacles.}
    \label{fig:overview}
\end{figure}

\begin{figure}[t]
\centering
    \includegraphics[width=\linewidth]{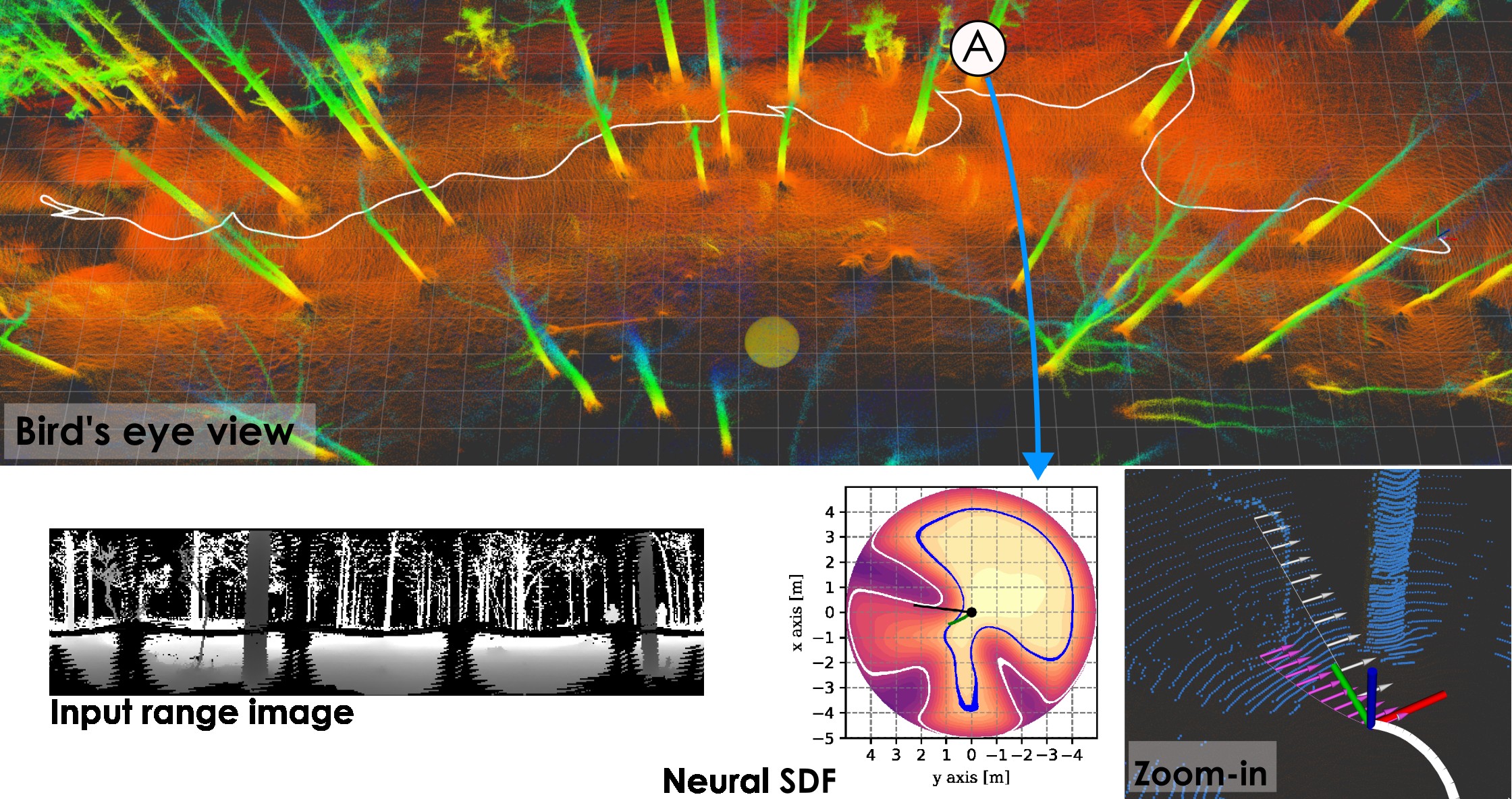}
    \caption{Visualization of a LiDAR-based flight among trees.
        The lower part of the figure depicts a zoom-in of a specific time instance,
        along with the corresponding LiDAR input range image, and the $z_B=0$ ``slice'' of the neural SDF encoding, with the obstacle line and the safety distance respectively in white and blue.}
    \label{fig:sdfnmpc}
\end{figure}

The NMPC is designed to track the reference trajectory provided by the exploration planner and provide acceleration commands. A neural network is used to process the sensor data directly and provide an SDF. This encoding is further used as a 3D position constraint in a local frame, while the reliance on only a local frame allows for improved resilience against odometry drift.
An overview block diagram of the method, denoted SDF-NMPC, is pictured in~\figref{fig:overview}.

First, the depth observations are encoded into a low-dimensional latent representation using a Variational Autoencoder~(VAE). This latent encoding is used as an input for the neural network approximating the SDF, which is trained in a supervised manner by sampling. This training is performed fully offline, and this neural architecture has been shown to efficiently generalize to new environments without online fine-tuning.

The neural network is embedded as a constraint into the NMPC architecture. In the case of a constrained field-of-view sensor, additional position constraints are enforced to ensure that the system evolves in the currently visible space (such that the sensor is always pointed in the direction of motion to capture potential obstacles).
Additionally, a theoretical analysis of the controller is proposed to derive recursive feasibility and stability conditions.

The SDF-NMPC was tested on the RMF in several forest environments to validate its collision-avoidance capability. During experiments, a human operator provided adversarial reference commands, intentionally trying to collide the robot with trees, branches, and bushes.
\figref{fig:sdfnmpc} shows a visualization of such a trajectory, using the LiDAR and without resorting to any mapping system. The spatial encoding of the currently visible environment is depicted with the color gradient. Darker areas represent obstacles and unseen space, and the robot is constrained to navigate within the blue line, ensuring collision avoidance. This method ensures safe navigation against visible obstacles even in the case of failure of the mapping system, allowing for a safe recovery.

The neural NMPC yields an acceleration setpoint. Further, redundant safety is achieved by filtering this setpoint with a Control Barrier Function (CBF)-based safety filter.
It is a computationally efficient reactive collision avoidance mechanism, synthesizing a composite CBF that describes safety with respect to the visible environment. It provides a formal safety certificate, which is used to filter out potentially unsafe actions from the higher-level controller.

\begin{figure}[t]
    \centering
    \includegraphics[width=0.8\linewidth]{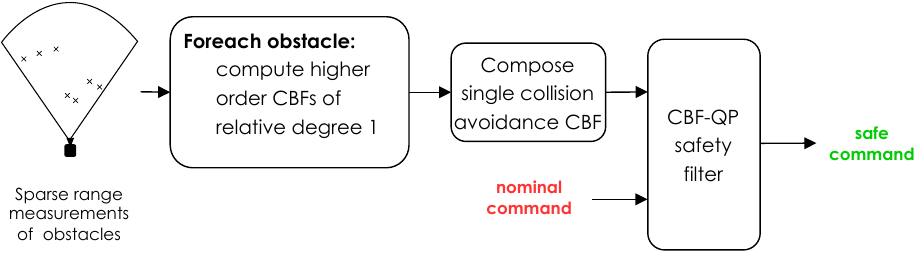}
    \caption{An overview of the proposed composite CBF safety filter.
        The method synthesizes a single, composite CBF describe the avoidance of several point-wise obstacles. This CBF is in turn used as the constraint of a QP problem, finding the minimal deviation from a nominal input that ensures a safe behavior.}
    \label{fig:cbf:overview}
\end{figure}

An overview block diagram is presented in~\figref{fig:cbf:overview}. From a conservatively sparsified point cloud measurement, a collection of high-order CBFs is computed, each describing safety with respect to the corresponding observed point. The composition method relies on computing an under-approximation of all those constraints. This continuous approximation yields a single, lumped constraint that describes safety with respect to all the visible obstacles.
While the safety filter requires solving a Quadratic Programming, an analytical solution is available, making the method extremely computationally efficient.

\subsection{Legged-Aerial Marsupial System}

\label{sec:marsupial}
\begin{figure}[t]
\centering
    \includegraphics[width=.24\linewidth]{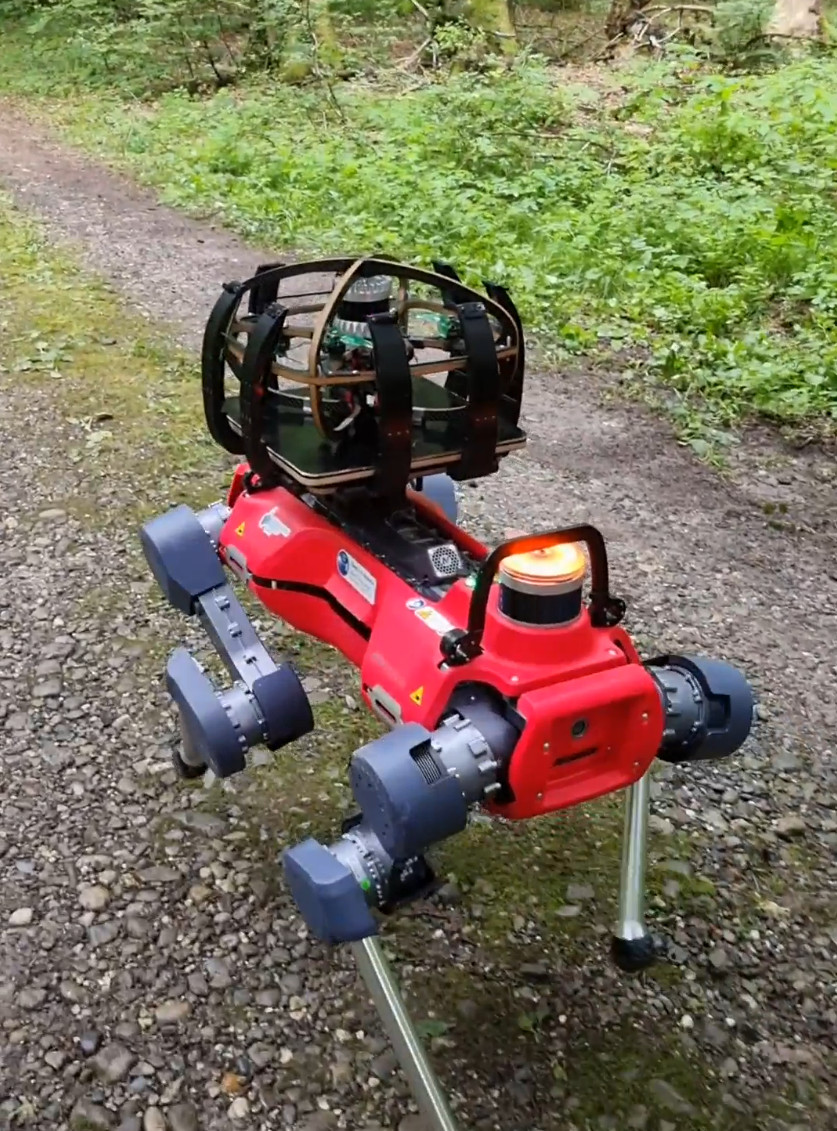}
    \includegraphics[width=.24\linewidth]{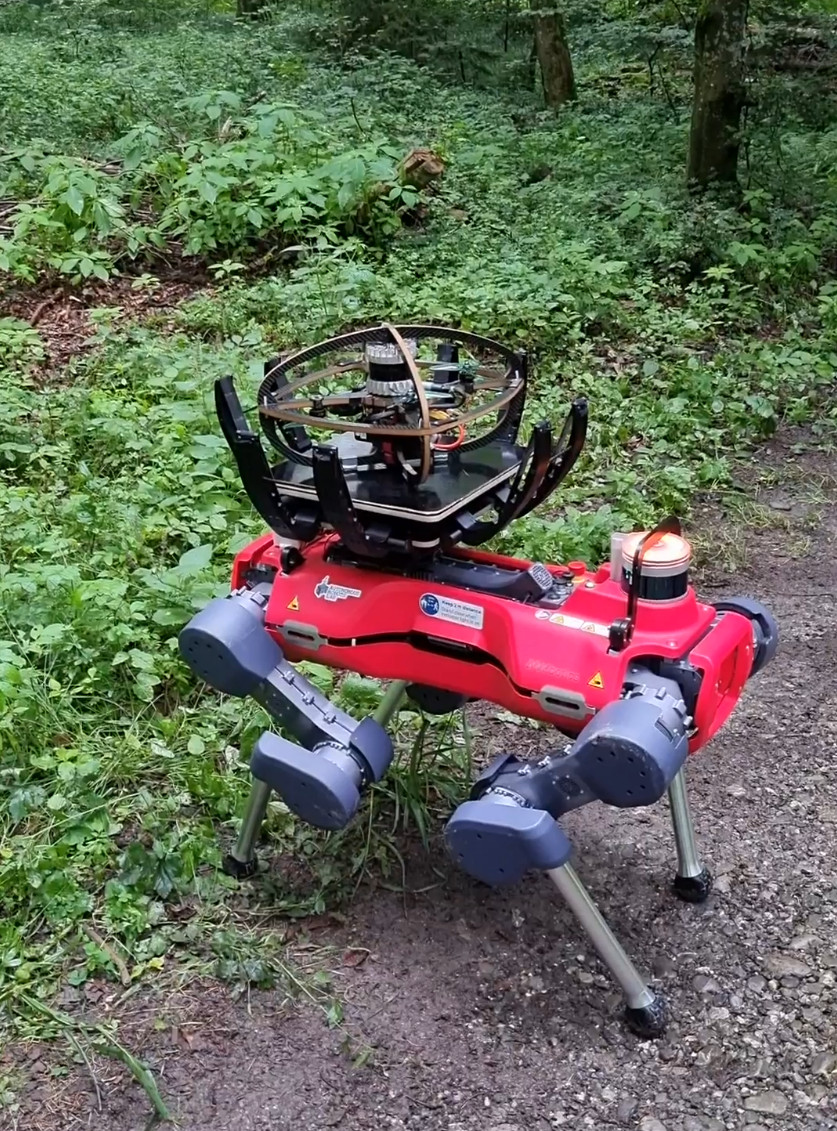}
    \includegraphics[width=.24\linewidth]{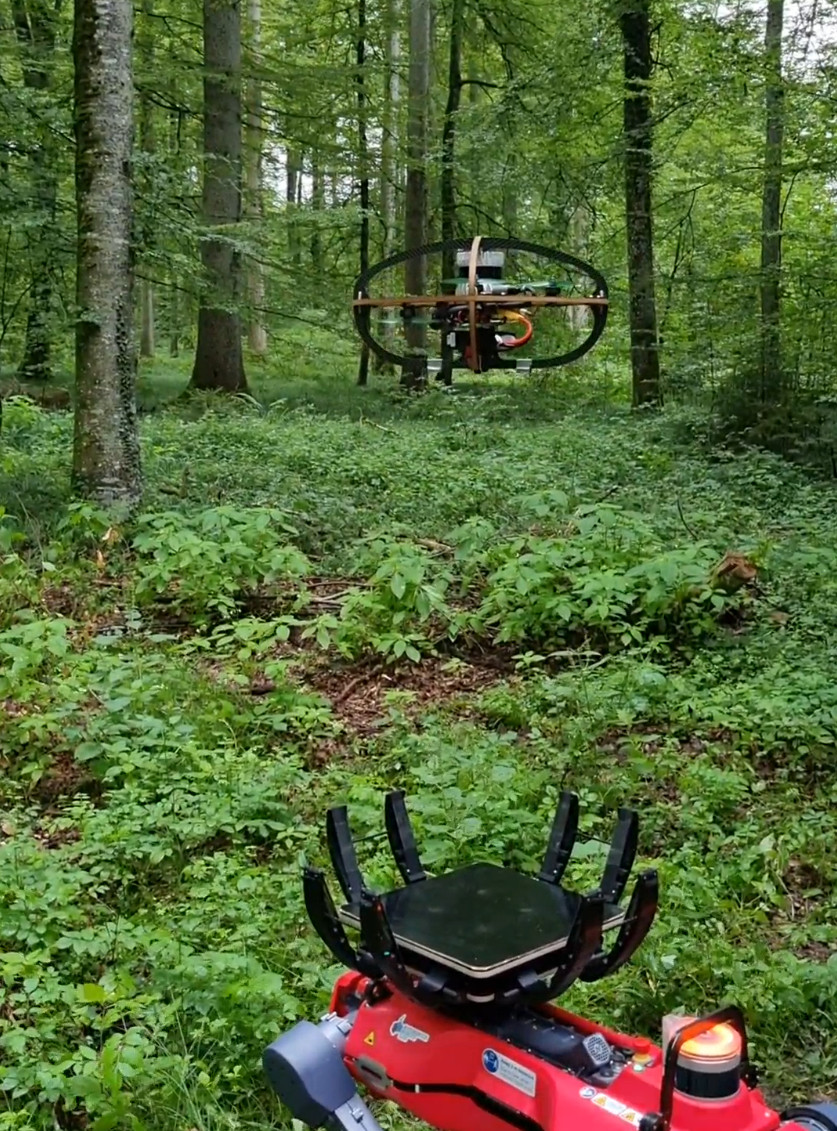}
    \caption{Deployment of a marsupial mission, including (left to right) the walking phase on easy terrain, the carrier mechanism opening, and the aerial robot starting the exploration mission.}
    \label{fig:marsupial_mission}
\end{figure}

A marsupial system is a multi-agent setup consisting of an aerial robot and a
quadruped. Combining the agility of the former with the endurance of the latter, this combination is
particularly well-suited for forestry exploration, facilitating
large-scale exploration while mitigating challenges related to difficult
terrain. We designed a carrier platform and mounted on the quadruped to transport the aerial robot during walking phases.
The carrier mechanism consists of two horizontal platforms and vertical
to prevent lateral motion. A single actuator, powered by the
battery of the quadruped, allow the mechanism to
open and the drone to take off (\figref{fig:marsupial_mission}).

The mechanism was validated by teleoperating the robot to an untraversable
area and deploying the flying robot, which followed the exploration plan detailed in Section \ref{sec:undercanopy-flight}.

Testing in forested environments of co-planning and map sharing method, such as the one detailed on our prior work \cite{zacharia2025ral}  is subject to future work. The method is able to assess whether a deployment of the aerial robot is necessary to perform the volumetric exploration (e.g., because of traversability issues) and, when the aerial robot is deployed, the two systems exchange a compressed shared map over wifi and individually perform mapping and exploration of the designed area.

\section{Automated Forest Inventory}
\label{sec:automated-forest-inventory}
In this section, we describe technologies for estimating tree traits and constructing forest inventories using maps acquired during the robot's navigation phase, either in real time (online) or post-mission (offline). We distinguish between three distinct processes:
\emph{Online inventory} generation occurs onboard one robot during a single navigation session, producing an inventory immediately usable by a forester (Sec.~\ref{subsec:forest-inventory})
\emph{Multi-Mission SLAM} involves the aggregation and processing of LiDAR data collected across multiple missions and different robots to create a longitudinal data acquisition via multi-mission LiDAR SLAM. We based our LiDAR SLAM method on \cite{oh2024iros}, which consistently reported up to 80\% of correct loop closures even for scans that are 5 m apart and 60\% when 10 m, effectively eliminating drift. The method was tested in four different forests in the UK, Finland and Switzerland.
\emph{Panoptic Segmentation} without real-time constraints, data provided by the Multi-Mission SLAM enables the use of more advanced and computationally intensive methods—such as panoptic segmentation for higher-fidelity analysis and interpretation of forest structure and tree traits (Sec.~\ref{sec:ubonn-dataset})

To support offline learning and evaluation of the last process, we collected a longitudinal dataset at three different time points. This dataset enables supervised training of learning-based algorithms for forest inventory tasks (Sec.~\ref{subsec:forest-panoptic-segmentation}).

\subsection{Online Forest Inventory}
\label{subsec:forest-inventory}
Despite being robot-agnostic, our real-time forest inventory pipeline was primarily tested and developed for the legged robot and is build upon the output of its state estimator (Sec. \ref{sec:forest-inventory-legged}), as described by Frei{\ss}muth et al.~\cite{Freissmuth2024}. This pipeline first performs online tree segmentation and reconstruction during mission execution by combining LiDAR-based SLAM, a Tree Manager module, and a Hough Transform-based modeling procedure. The Tree Manager module applies an cloth simulation filtering to remove the ground from the SLAM payloads, then the normalized terrain is segmented out and DBSCAN and Non-Maxima Suppression are applied to the filtered cloud to distinguish tree trunk cylinders. A spatio-temporal aggregation then is used to  incrementally build globally consistent maps and extract tree-level traits such as diameter at breast height (DBH) with accuracy comparable to terrestrial laser scanning, while operating without post-processing. The system was deployed on the ANYmal quadruped robot within the software stack described in Section \ref{sec:forest-inventory-legged} and validated in operational forest environments.

\begin{figure}
 \centering
 \includegraphics[height=3.8cm]{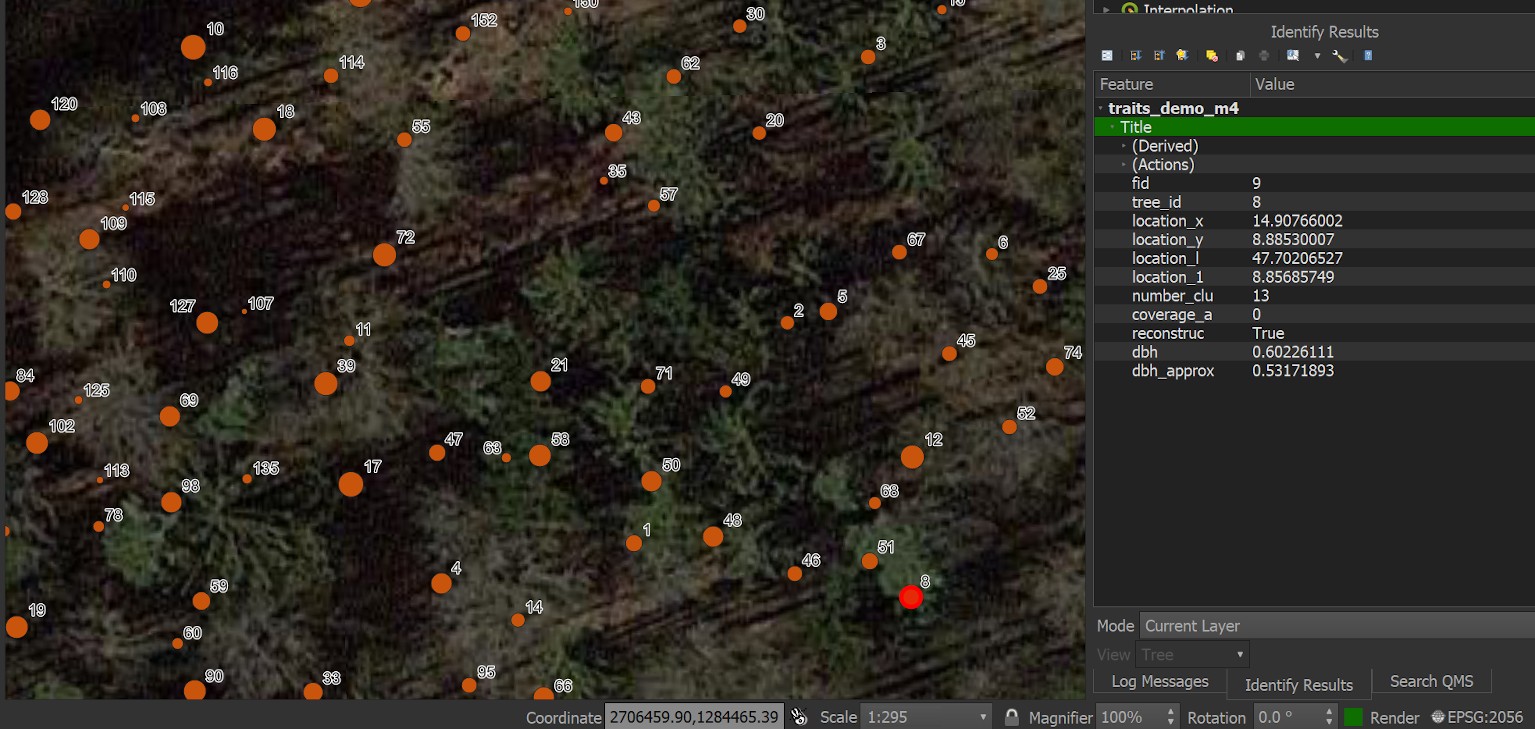}
 \includegraphics[height=3.8cm]{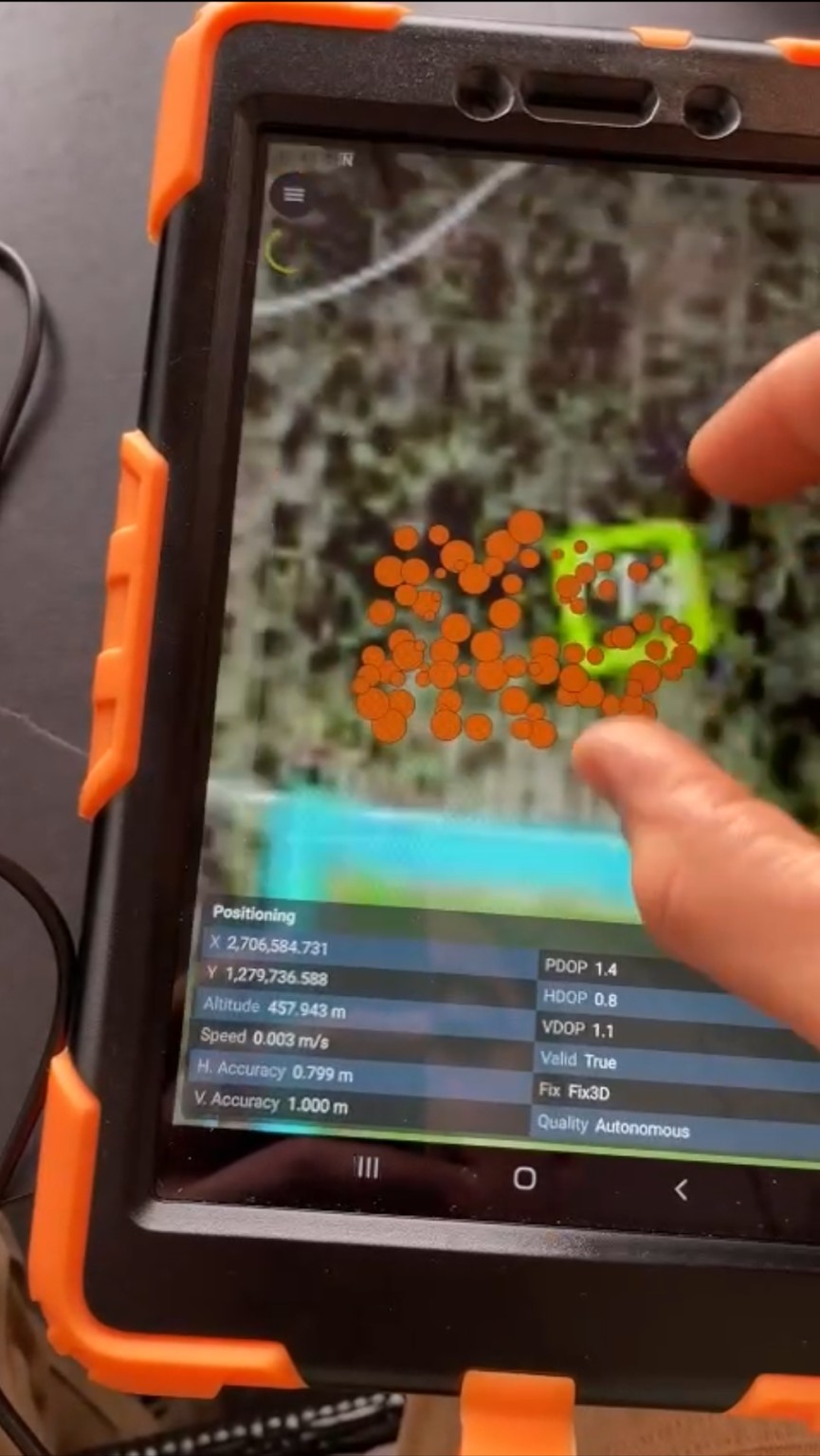}
 \caption{Integration of the spreadsheet output from the online forest
inventory \cite{Freissmuth2024} onto the QGIS software on PC (left) and tablet
(right)}
  \label{fig:qgis-integration}
\end{figure}

The outputs from the online forest inventory, available both as structured tabular files and through a dedicated SLAM interface, were integrated into QGIS, a widely used open-source Geographic Information System, allowing forestry researchers and practitioners to explore tree-level data and integrate it into their workflow  (\figref{fig:qgis-integration}).

\subsection{Longitudinal LiDAR Dataset}
\label{sec:ubonn-dataset}
To facilitate training of semantic scene segmentation approaches for forestry applications, we collected and densely labeled a longitudinal dataset for forestry robotics called ``DigiForests'', which also provides the foundation to assess the performance of forestry scene interpretation methods.
We provide here a brief overview of the dataset and refer to our dataset paper~\cite{malladi2025icra} and website\footnote{The dataset is available
at \url{https://digiforest.eu/datasets}.} for further details.

The dataset was collected in Oberwald forest, see \figref{fig:stein-map}, in three different campaigns: March 2023, October 2023, and July 2024. Therefore, our dataset provides data from different growth periods with varying amounts of shrub and under-canopy vegetation, which makes downstream tasks, like DBH estimation, potentially challenging due to the reduced visibility of the trunks.

The plots in the dataset show different levels of complexity in terms of traversability and clutter, but also differ in forest types, species, and silvicultural management regimes. For reference, a manually-built forest inventory includes measurements of tree DBH, length of clear wood, height of first green branch, and tree species.

To allow coverage of a large forest area, but also capture trunks and below-canopy scans at the same time, we used a different sensor setup employing similar LiDAR sensors used on the robotic platforms.
Using a backpack carried by a human allows reaching challenging areas in the forest that are currently not reachable by legged robots, but also difficult for drones due to dense foliage.
Therefore, our backpack solution is more practical for acquiring large amounts of LiDAR data in challenging scenarios with fewer restrictions on traversability or forest types, while featuring similar sensor configurations employed on the different robotic platforms. Differences in the positioning of the sensor, but also the number of LiDAR beams, are less relevant in the application scenario, where aggregated point clouds are used for a more fine-grained analysis as described in Sec.~\ref{subsec:forest-panoptic-segmentation}.

In particular, we used a backpack-mounted sensor package consisting of a Hesai XT32 LiDAR inclined at 45 degrees (March 2023), and a Hesai QT64 horizontally aligned (October 2023).
In July 2024, we collected data with a combination of a horizontally aligned  Hesai QT32 and a Hesai XT32 inclined at 45 degrees, to ensure maximum coverage of the tree stems and tree canopy.
The sensor package was additionally equipped with a GNSS receiver for geo-referencing the data, and multiple cameras were employed for visual odometry.

Additionally, for the March 2023 and July 2024 campaigns, we used a UAV to record above-canopy aerial data covering the entire forest region shown in~\figref{fig:stein-map}. For the
March 2023 campaign, we flew a DJI Matrice M600 equipped with a sideways-mounted
Velodyne HDL-32E while for the July 2024 campaign a Hesai XT-32M2X was used. We used on-board RTK GPS to globally geo-reference the drone via GNSS poses to register the corresponding backpack sensor point clouds from each plot.

The data was furthermore post-processed using SLAM techniques to generate aggregated point cloud maps.
A key achievement is the spatial alignment of recorded data
across seasons.
To achieve sequence-to-sequence alignment across different data collection seasons, we jointly optimized a pose graph over an individual plot's trajectories from all sessions, globally aligning all point clouds spatially~\cite{casseau2024iros}.

\begin{figure}[t]
  \centering
  \begin{subfigure}[t]{0.32\textwidth}
    \centering
    \includegraphics[width=\textwidth]{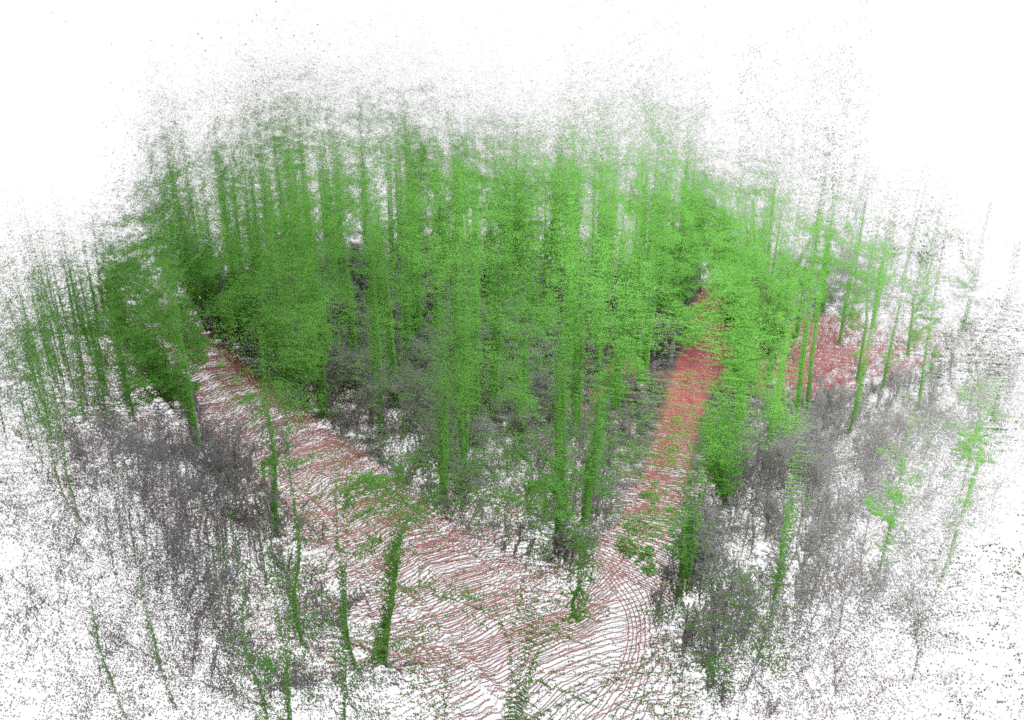}
    \caption{}
    \label{fig:tree_semantics}
  \end{subfigure}
  \hfill
  \begin{subfigure}[t]{0.32\textwidth}
    \centering
    \includegraphics[width=\textwidth]{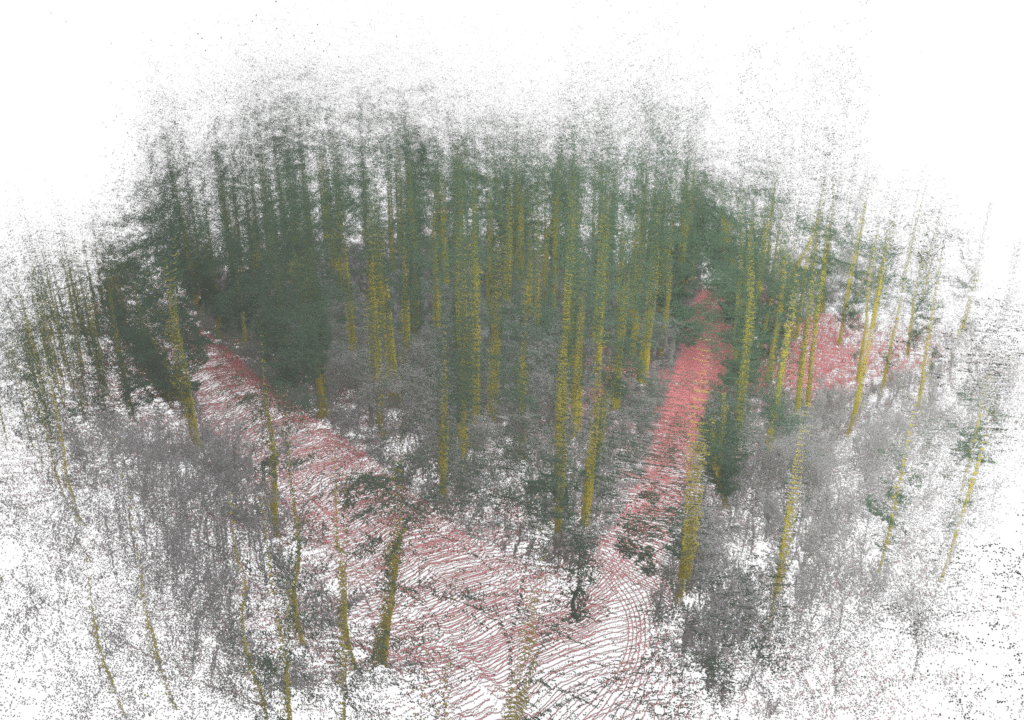}
    \caption{}
    \label{fig:tree_stem}
  \end{subfigure}
  \hfill
  \begin{subfigure}[t]{0.32\textwidth}
    \centering
    \includegraphics[width=\textwidth]{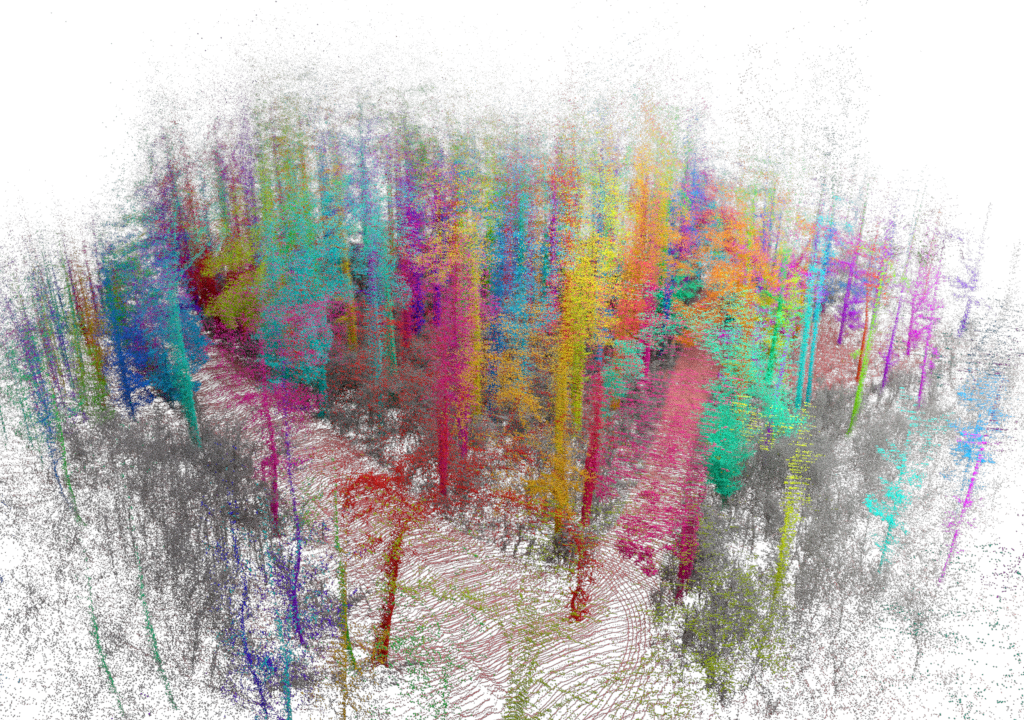}
    \caption{}
    \label{fig:tree_instance}
  \end{subfigure}

  \caption{The ``DigiForests'' dataset provides a longitudinal dataset of multiple forest plots recorded with a mobile LiDAR scanning system.
    The data has been recorded in three different seasons: Spring, Autumn, and Summer, with very different appearances in terms of vegetation.
    Besides a semantic annotation of the data into \textcolor{treecolor}{tree}, \textcolor{shrubcolor}{shrub}, and \textcolor{groundcolor}{ground} shown in~(a), we also provide fine-grained annotations of \textcolor{stemcolor}{stem} and \textcolor{ForestGreen}{canopy} of trees shown in~(b), and instance annotations for trees shown by different colors in~(c).}
  \label{fig:motivation}
\end{figure}

We manually labeled and verified all recording sequences using an extension of the point cloud annotation tool developed originally for the SemanticKITTI dataset~\cite{behley2019iccv}.
Examples of the annotated data are shown in \figref{fig:motivation}.

\subsection{LiDAR-based Forest Panoptic Segmentation}
\label{subsec:forest-panoptic-segmentation}

For a more fine-grained analysis of the gathered LiDAR point cloud, we employ a panoptic segmentation approach to add the required information for downstream tasks, such as deriving tree traits, which are relevant for providing the decision support system for forestry management (see Sec.~\ref{sec:dss}).
Note that the offline processing allows for processing aggregated scans and additional post-processing that is not possible in an online setting.
Furthermore, more computationally demanding approaches can be employed, which lead to a more complete forest inventory.

Our approach to panoptic segmentation builds on commonly employed semantic and instance segmentation for urban and agricultural environments.
However, applying these techniques directly to forestry presents unique challenges due to the hierarchical and overlapping nature of the annotations, especially for the tree, i.e., stem and canopy classes, described in Sec.~\ref{sec:ubonn-dataset}.
To effectively leverage such approaches for forest scene interpretation, it is necessary to adapt these methods to handle the specific structure and requirements of our application.

To address these challenges, we designed a panoptic segmentation approach tailored to forest scenes captured by LiDAR sensors.
For segmentation, we use Minkowski Engine~\cite{choy2019cvpr}, an auto-differentiation library for sparse tensors suitable for efficiently processing sparse 3D point clouds, which is crucial for handling the large, mostly empty spaces typical of LiDAR data.
This leads to significant savings in both memory and computational resources compared to dense convolutional neural networks using voxel grids.

\begin{figure}[t]
	\centering
  \includegraphics[width=.85\linewidth]{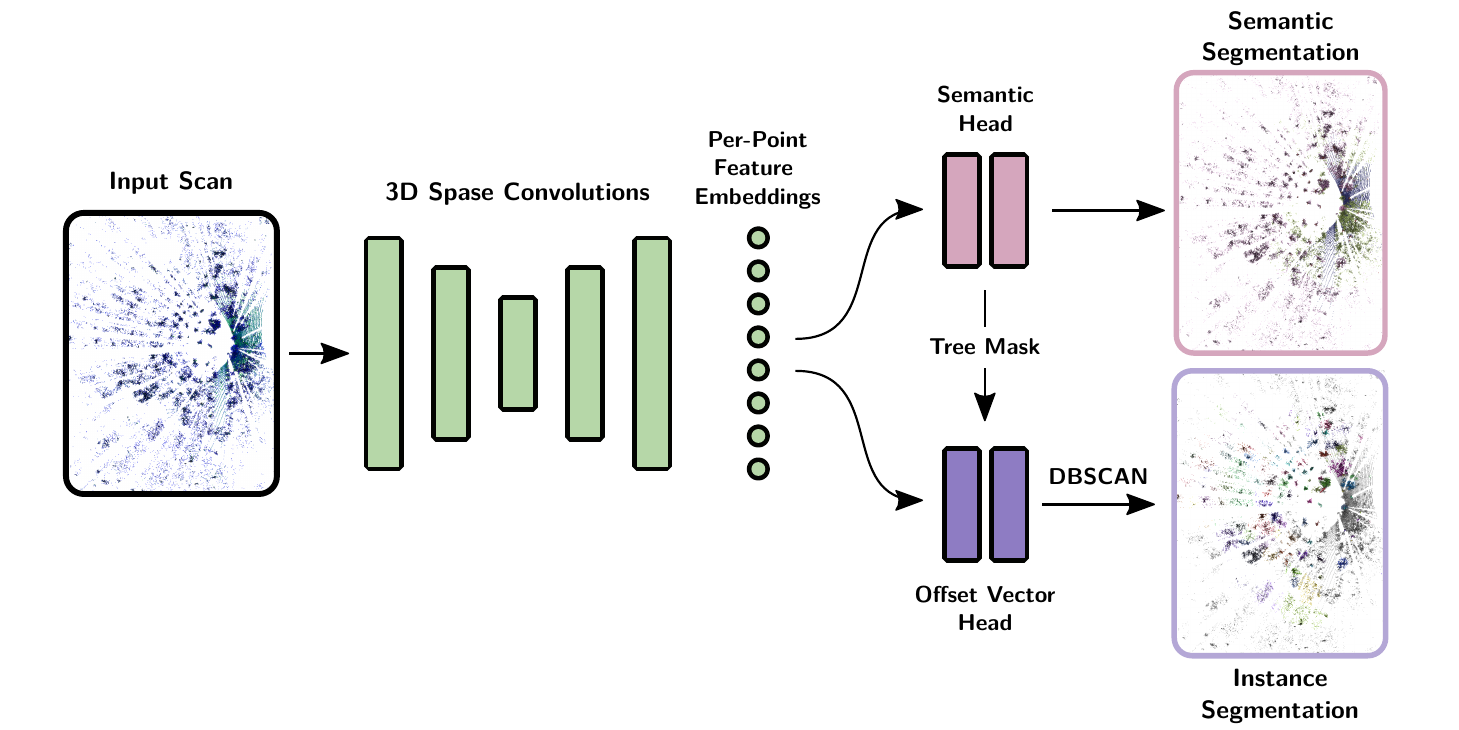}
	\caption{Overview of our panoptic segmentation approach. We use a LiDAR scan as input for a sparse 3D convolutional encoder-decoder that generates a feature embedding per LiDAR point. The feature embeddings are passed to two multi-layer perceptron (MLP) heads, resulting in per-point semantics from the semantic head and instance segmentation from the offset vector head via clustering via DBSCAN.}
	\label{fig:networkarch}
\end{figure}

The network architecture of our approach is shown in \figref{fig:networkarch}.
We use a MinkUNet architecture~\cite{choy2019cvpr}, which is an extension of the U-Net~\cite{ronneberger2015micc}
model specifically optimized for sparse data using the Minkowski Engine.
The encoder-decoder architecture produces a feature embedding vector of size 32 for each point in the input LiDAR scan.
The feature embedding is then passed on to two network heads in parallel: a semantic head responsible for estimating the semantic classes and an offset vector head predicting offsets that can be used to derive instances. More specifically, the semantic segmentation head is a two-layer multi-layer perceptron~(MLP) that predicts per-point probabilities over the classes: ground, shrub, stem, or canopy. Moreover, the offset vector head is a four-layer MLP  predicting per-point three-dimensional offset vectors, which point to the geometric center of a potential tree instance.

For deriving individual tree instances, we first determine points classified as stem or canopy and then apply the predicted three-dimensional offset vectors to these tree points, shifting them towards the estimated center.
The shifted points are then clustered using the DBSCAN~\cite{ester1996kdd} to group them in individual tree instances.  More details on the approach can be found in our dataset paper~\cite{malladi2025icra}.
Our method performs both semantic segmentation and tree instance segmentation, addressing the unique hierarchical labeling structure presented in Sec.~\ref{sec:ubonn-dataset}.

\section{Decision Support System}
\label{sec:dss}
\begin{figure}
    \centering
    \includegraphics[width=0.9\linewidth]{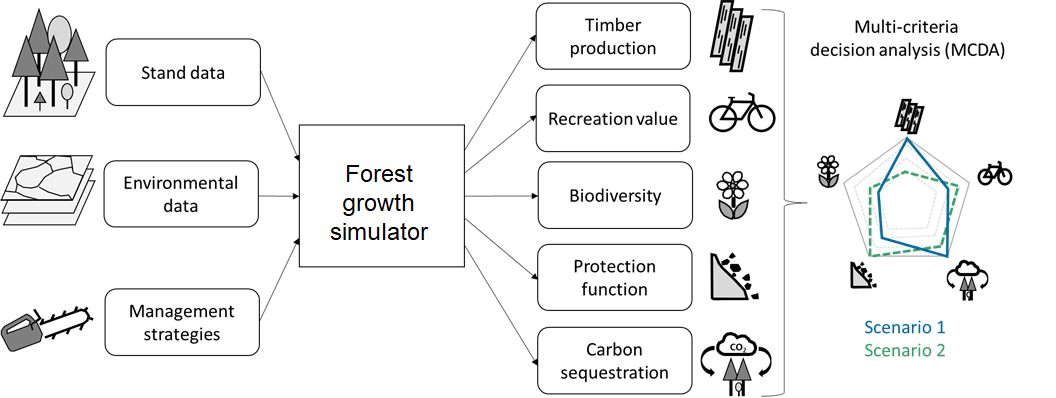}
    \caption{Overview of the Decision Support System pipeline (adapted from Thrippleton/WSL).}
    \label{fig:dss-overview}
\end{figure}

The Decision Support System (DSS) combines the robot-derived forest inventory with environmental data and management rules to support interventions at two horizons. In the short term, stand- and tree-level attributes (e.g., DBH, height, form, biomass/volume proxies, health) underpin choices about which trees to remove or retain and the expected economic return. In the long run, growth simulators model stand development under various climate scenarios and management paths, enabling multi-objective trade-offs between carbon sequestration, resilience, timber yield, protection, and recreation. (Fig.~\ref{fig:dss-overview}).

As an initial case study, we used the climate-sensitive gap model ForClimV4.1 \cite{Bugmann} for this kind of future forest growth estimation. We initialized it using manually collected ground truth reference data, including information about Diameter at Breast Height (DBH), tree species, and current health status of a small forest patch within the Stein am Rhein study area.

\begin{figure}
 \centering
 \includegraphics[width=\textwidth]{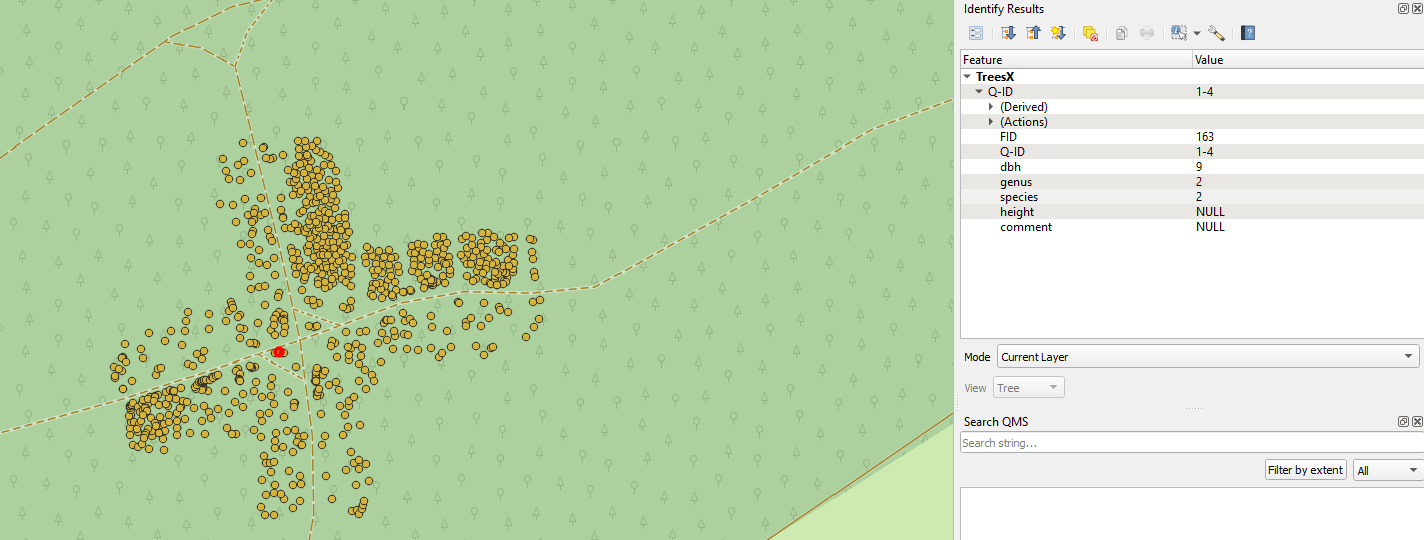}
 \caption{Individual tree data collected by team-WSL at Stein am Rhein, Switzerland as seen from the data inventory software.}
 \label{fig:wsl-data-collection-map}
\end{figure}

In total, the individual tree data collection included 800 trees, covering a wide variety of forest types (e.g., conifers/broadleaves; dense and light stands; continuous cover and rotation forestry) and will be made publicly available at a later stage. A snippet of the forest area map of the data collected is shown in \figref{fig:wsl-data-collection-map}.

We formulated potential forest management strategies together with the forester of the site and implemented them into the growth simulator. Simulations have been conducted for a time horizon of ca. 100 years, thereby considering historical climate data as well as different estimated climate scenarios (RCP4.5 and RCP8.5). The results will be coupled to indicator value functions allowing the calculation of future provision of biodiversity and ecosystem services. In future work, we intend to extend this approach to use data collected from robotic systems and extract individual tree parameters and process in the tree growth simulator.

\section{Autonomous Harvesting}
\label{sec:harvester}

The SAHA robot is designed to perform cutting and forwarding operations at the same time, minimizing the number of steps required in the harvesting process.
The lightweight design of the robot minimizes the impact on the forest floor and allows it to traverse and operate in confined spaces, which is crucial for selective thinning.
An active suspension chassis with four independently actuated legs provides stability on uneven ground.
The chassis features an articulated steering mechanism, allowing relative rotation between the front and back parts of the vehicle~\cite{Corke2001}.
The arm offers six degrees of freedom, of which four are hydraulically actuated, and carries a bidirectional harvester head.

A compact perception module combining a LiDAR, an RGBD camera, an IMU, and a GNSS receiver is mounted on the side of SAHA's arm, enabling localization, navigation, and tree detection.
Onboard compute is provided by an NVIDIA Jetson Orin module integrated in the perception module and an industrial computer performing real-time control.
A pipeline for autonomous operation of SAHA has been developed, with a number of its modules tested in the field, demonstrating fully automated cutting operations of target trees specified in the prebuilt map.

The following sections describe the main components of the autonomous harvester and their integration into a complete system.

\subsection{Control Automation}
Various controllers have been implemented for the robot, providing key functionalities such as terrain adaptation, driving control, and arm control.
The terrain adaptation controller is adapted from our previous implementation~\cite{hutter2016force} and
 consists of a cylinder force controller and a chassis pose controller.
The chassis pose controller utilizes a Virtual Model Controller (VMC) to compute desired force and torque on the chassis.
Then, hierarchical optimization is used to distribute the support forces of the legs to maintain stable contact when the robot operates on uneven forest ground.
The desired support forces from individual legs are tracked by the cylinder force controller, which relies on the feedback from the chassis cylinder pressure measurements.
For more details of the implementation, we refer the readers to our previous work~\cite{hutter2016force}.

\begin{figure}
    \centering
    \begin{subfigure}{0.245\linewidth}
        \includegraphics[width=\linewidth]{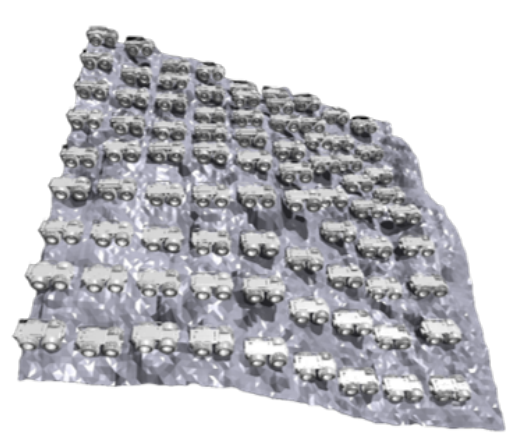}
        \caption{Massively spawned SAHA on randomly generated terrain}
    \end{subfigure}
    \begin{subfigure}{0.245\linewidth}
        \includegraphics[width=\linewidth]{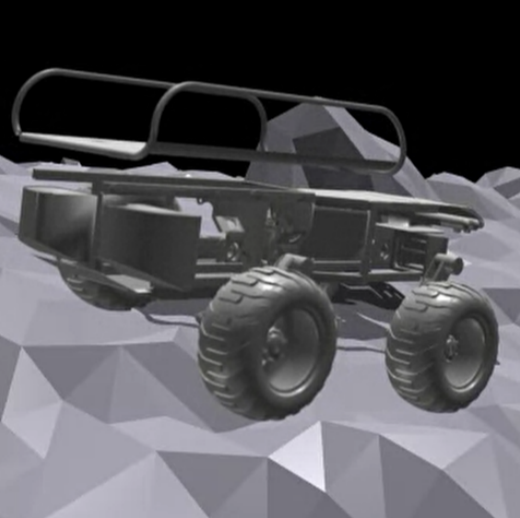}
        \caption{SAHA balancing on a traversable terrain}
    \end{subfigure}
    \begin{subfigure}{0.245\linewidth}
        \includegraphics[width=\linewidth]{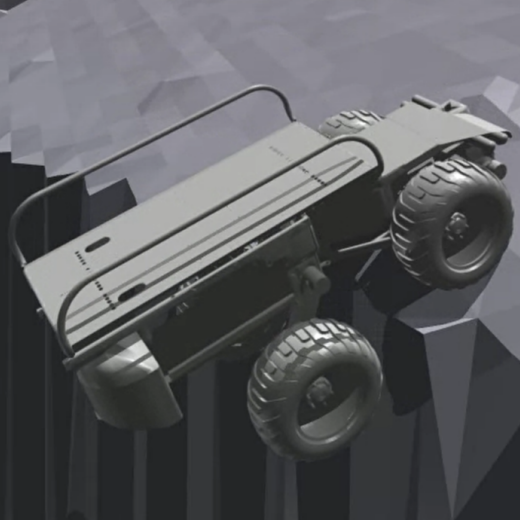}
        \caption{SAHA fails to balance on a steep step}
    \end{subfigure}
    \begin{subfigure}{0.245\linewidth}
        \includegraphics[width=\linewidth]{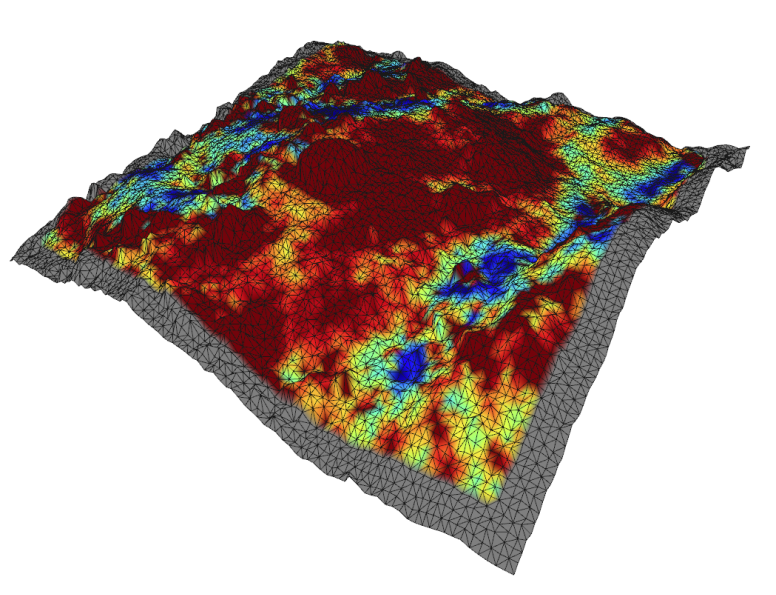}
        \caption{Traversibility cost map}
    \end{subfigure}
    \caption{SAHA traversability evaluation in parallel simulation.}
 \label{fig:SAHA_terrain_traversibility}
\end{figure}

While this optimization-based approach aims to find the most stable leg configuration subject to SAHA's kinematic limits, there exist difficult terrain patches that the balancing controller would fail to maintain stable ground contact.
To quantify the limit of the balancing controller, we leverage massive parallel simulations.
As seen in Figure~\ref{fig:SAHA_terrain_traversibility}, we spawn several SAHA robots on a randomly generated terrain.
Each robot attempts to establish a stable ground contact.
The terrain patches under the successful ones are marked as traversable and receive a traversability score based on the remaining margin for chassis pose control, while the others are marked as not traversable.
Based on the simulation result, we use the result to train a geometric-based traversability prediction module similar to~\cite{Frey2022Traversability} and use it as the traversability cost in local navigation

Assuming the terrain adaptation controller handles the terrain uncertainties, the driving controller is designed based on a planar simplification of the kinematic model of the center-articulated chassis of SAHA.
The driving controller enables SAHA to reliably track given reference targets.
The derivation and convergence analysis is presented in~\cite{hu2024motion}.

Once SAHA arrives in the vicinity of the target tree, the arm must be accurately controlled to move the harvester head to the target tree trunk.
While controlling hydraulically actuated robot arms is complex, we use a learning-based controller with online adaptation capabilities~\cite{nan2024learning} to adapt to SAHA's hydraulic component dynamics at test time.
As reported in~\cite{nan2024learning}, it reaches around \SI{5}{\centi\meter} accuracy for end effector trajectory tracking, which provided adequate performance for harvesting.

\subsection{Navigation in Scanned Forest}
To autonomously navigate to the selected tree, we utilized the LiDAR map of the target area from the previous scans of the legged and aerial platforms.
The SAHA robot localizes itself within the prebuilt map frame by matching the onboard LiDAR point cloud with the one from the map.
Upon receiving a cutting target, a series of waypoints is generated between SAHA's current location and the target tree.

\begin{figure}
    \centering
    \begin{subfigure}[b]{0.49\linewidth}
        \includegraphics[trim={3cm 2cm 4cm 1.5cm},clip,width=\linewidth]{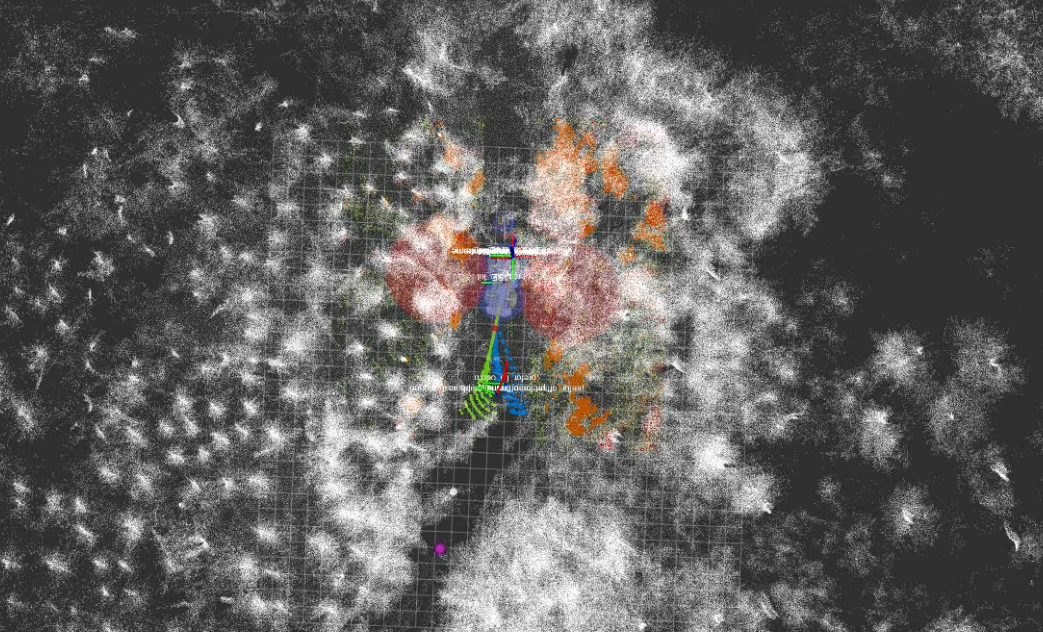}
    \end{subfigure}
    \begin{subfigure}[b]{0.486\linewidth}
        \includegraphics[width=\linewidth]{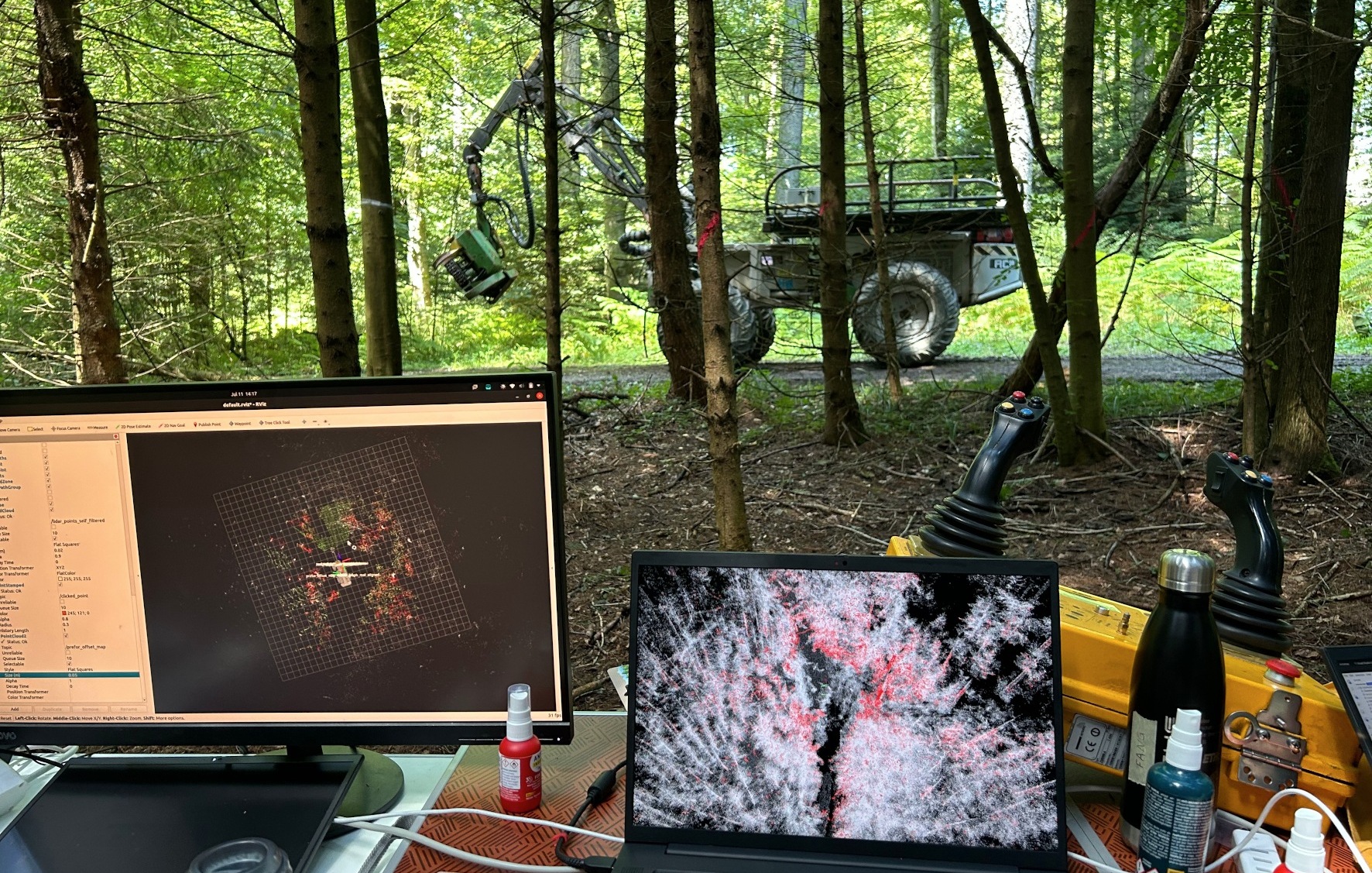}
    \end{subfigure}
    \caption{SAHA navigating through the forest with a prior point cloud map. (Top) Prebuilt point cloud map in white, matched local point cloud in orange. Navigation target in purple, and motion primitives selected by local planner in green. (Bottom) picture of SAHA during autonomous navigation.}
 \label{fig:SAHA-navigation}
\end{figure}

The waypoints are then used by a local path planner to generate a local path considering obstacles and traversability.
Utilizing prebuilt motion primitives with offline simulation, the local navigation module selects collision-free motion plans that can be effectively tracked by SAHA's unique articulated steering system.
During local planning, a real-time terrain analysis module generates traversability labels of the area surrounding SAHA based on geometric data.
The traversability cost is considered by the local planner to simultaneously minimize risk while selecting the best primitive in the target direction.
This capability ensures safe and efficient navigation in the complex and unpredictable forest environment. For more details on the local navigation algorithm, see \cite{hu2024motion}.
We deployed the Traversability-Aware Navigation algorithm detailed in~\cite{hu2024motion} in the Stein am Rhein forest.
A picture of the navigation system running on the SAHA robot is shown in \ref{fig:SAHA-navigation}.

\subsection{Map-Based Tree Grasping and Cutting}
\begin{figure}
 \centering
 \includegraphics[width=.5\textwidth]{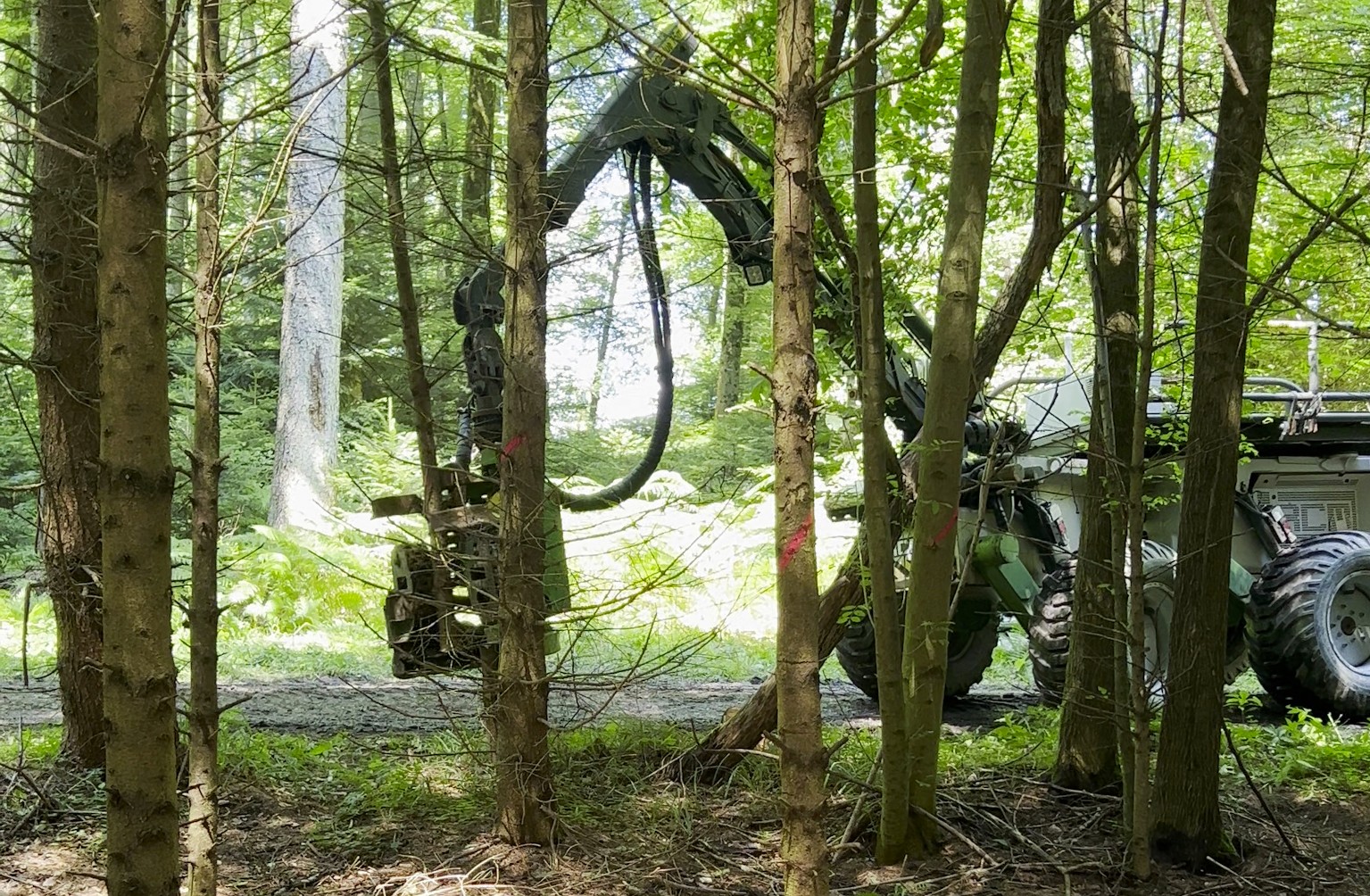}
 \caption{The SAHA robot releases the cut tree after manual triggering.}
 \label{fig:harvester-cut}
\end{figure}

Using the prebuilt map, the user can specify the target tree location through a
simple RViz interface. The state machine first evaluates the distance between
the robot's current position and the specified tree to determine whether it is
within the arm's reachable range. If the tree is out of reach, the autonomous
navigation module is triggered to drive the robot closer to the specified
location. Once the tree is within the arm's reachable range, the system
activates SAHA's automatic cutting mode. The arm is then controlled to reach
the target trunk, and once the clamp is in position, the operator is required to
manually trigger the button to initiate cutting (Figure
\ref{fig:harvester-cut}), ensuring safety.

\section{Experimental Evaluation}
\label{sec:experimental}
\begin{figure}[bt]
 \centering
 \includegraphics[width=.9\textwidth]{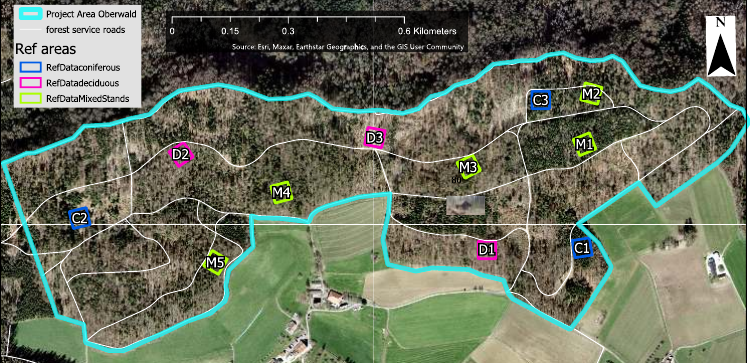}
 \caption{Map of the Stein am Rhein test site plots. The area is divided into labeled plots based on dominant tree type: \DeciduousSquare{} deciduous (D), \ConiferousSquare{} coniferous (C), and \MixedSquare{} mixed-species (M).}
 \label{fig:stein-map}
\end{figure}

The technologies presented by DigiForest have been extensively tested in real-world scenarios in Finland, the UK, and Switzerland. One of the principal test sites used is the Oberwald forest near Stein am Rhein, Switzerland. The site represents a characteristic Central European forest environment with deciduous, coniferous, and mixed stands. The area is actively managed and provides suitable conditions for testing autonomous systems under realistic operational constraints, including varied terrain, dense vegetation, and canopy occlusion.
For experimental purposes, the forest has been subdivided into plots based on dominant tree type: ``D'' for deciduous, ``C'' for coniferous, and ``M'' for mixed-species stands, as shown in \figref{fig:stein-map}

\subsection{Autonomous Legged Navigation}
\begin{figure}[h]
    \centering
    \includegraphics[width=0.49\linewidth]{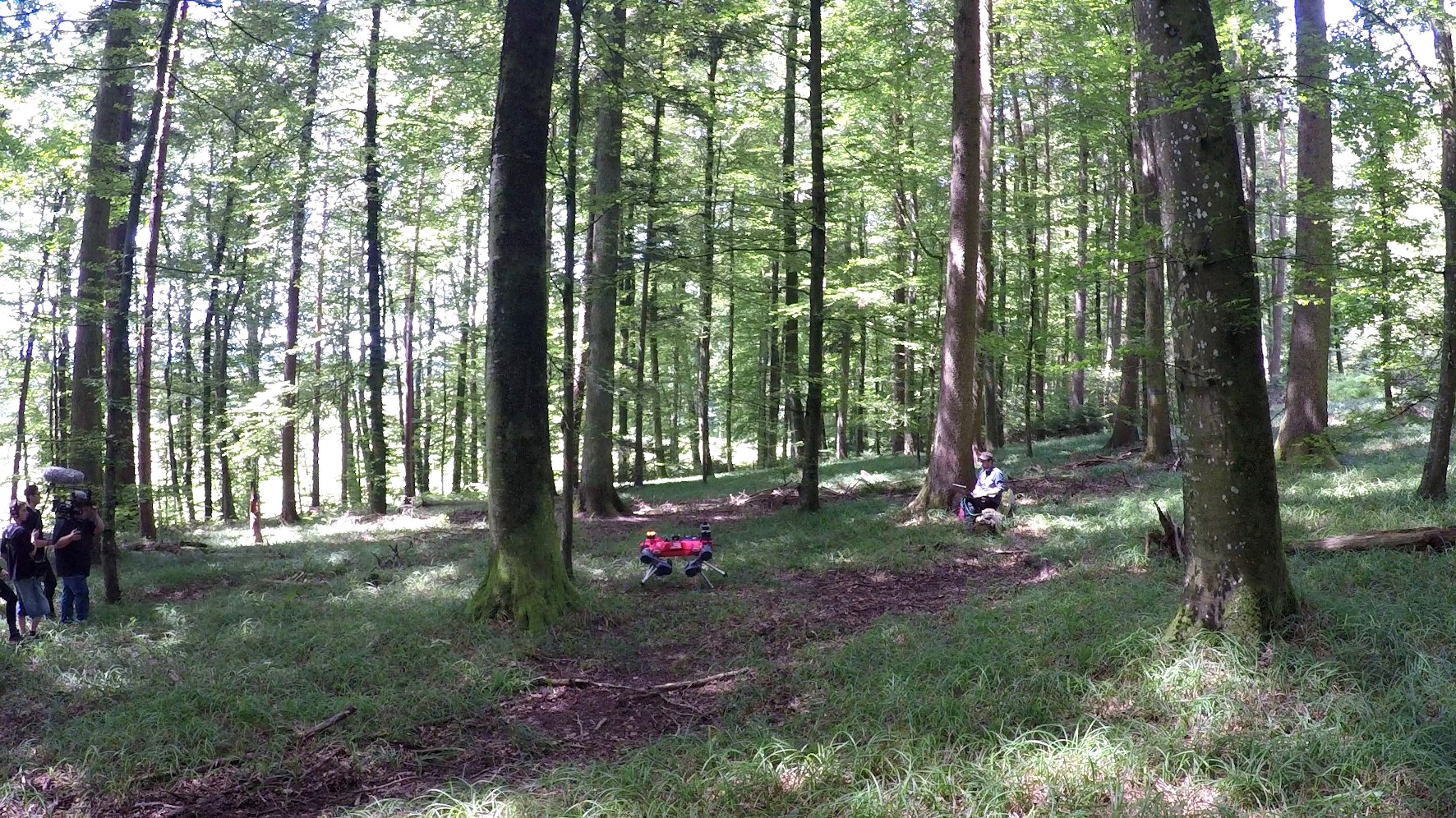}
    \includegraphics[width=0.49\linewidth]{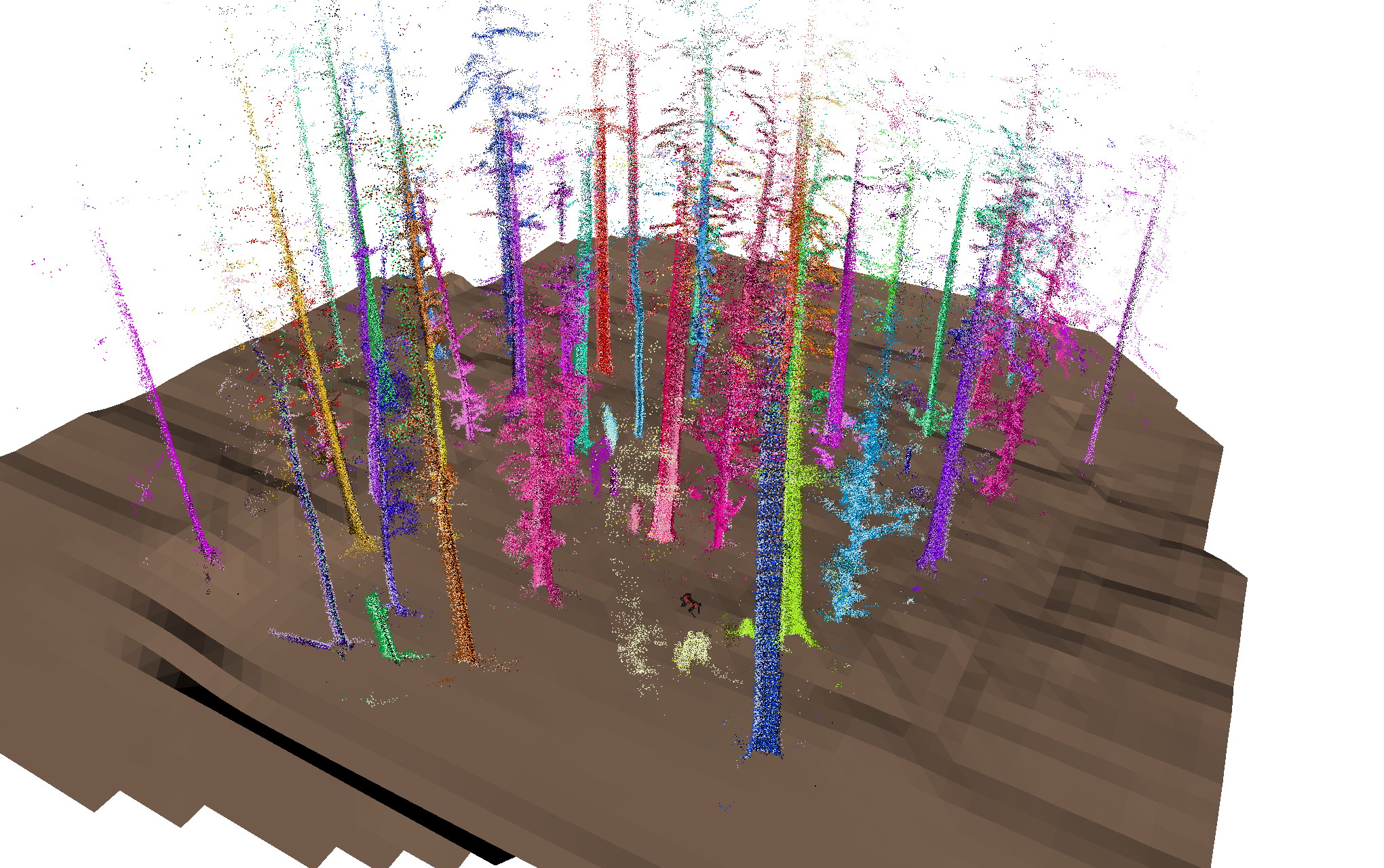}
    \caption{Autonomous deployment of ANYmal in Oberwald: \emph{Left:} ANYmal surveying the \emph{M4} plot, with mixed trees. \emph{Right:} Tree segmentation and digital terrain model after the mission.}
    \label{fig:anymal-deployment}
\end{figure}

We deployed the quadruped robot in plot M4 of the Oberwald forest, as shown in Fig.~\ref{fig:anymal-deployment}. The plot was chosen because it presented less undergrowth and high grass than other plots in the forest, which were more challenging for the robot to traverse due to its comparable height ($\sim$\SI{60}{\centi\meter}). The ANYmal robot was deployed autonomously four times in the same plot, with each mission taking approximately 8 min. On each run, our forest inventory system was executed while the robot navigated autonomously in the field. We report \SI{102.5}{\meter} average for the Mean Distance Between Interventions (MDBI) metric for a path of \SI{152}{\meter}, indicating the robot operated autonomously for most of the mission. The sequences covered areas of up to \SI{0.2}{\hectare}, consistently reporting around 40 trees on each mission, which is a sign of the repeatability of the forest inventory system.

The quality of the robot trajectory as well as the map estimated by a SLAM system depends on obtaining frequent loop closures. This is challenging in forest environments due to its self-repeating geometry as well as strong occlusions in sensor observations. In the work by Oh et al.~\cite{oh2024iros}, we analyze the capabilities several state-of-the-art LiDAR place recognition systems, and develop a pipeline for obtaining robust loop-closures with baselines ranging from \SIrange{5}{10}{\meter}. The pipeline provides reliable loop-closures by leveraging a candidate proposal models learned on forestry datasets~\cite{vidanapathirana2022icra}, followed by a geometric verification step and a final refinement using Iterative Closest Point (ICP) registration.

 \subsection{Autonomous Undercanopy Flight}
 The autonomous exploration stack was tested in Oberwald with two depth sensing modalities: LiDAR on the BLK2FLY and depth from the MRL's RealSense D455.

\begin{figure}[t]
	\centering
	\includegraphics[width=.7\columnwidth]{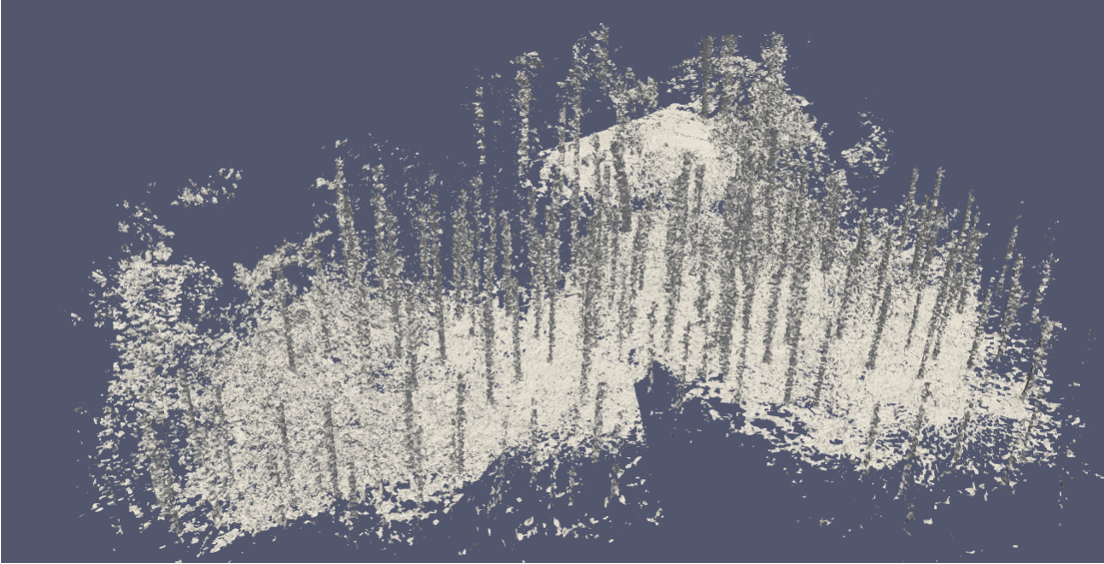}
	\caption{3D mesh reconstructed by BLK2FLY in Stein am Rhein.}
	\label{fig:blk_reconstruction}
\end{figure}

Since access to the computing unit onboard the BLK2FLY was unavailable, we streamed the data to a laptop in the field, where all computations on the exploration stack were performed. Whenever a new path was planned, it was sent to the BLK2FLY via a ROS interface, and the UAV continued with a regular exploration mission. An example of the 3D reconstruction obtained with the BLK2FLY from a test site in Stein am Rhein is shown in \figref{fig:blk_reconstruction}.

We also demonstrated our exploration algorithm with the MRL UAV. Two major tests were performed: demonstrating the benefits of performing Visual-Inertial SLAM with submap-to-submap alignment and exploration in the forest.

\begin{figure*}
	\centering
	\includegraphics[width=\linewidth]{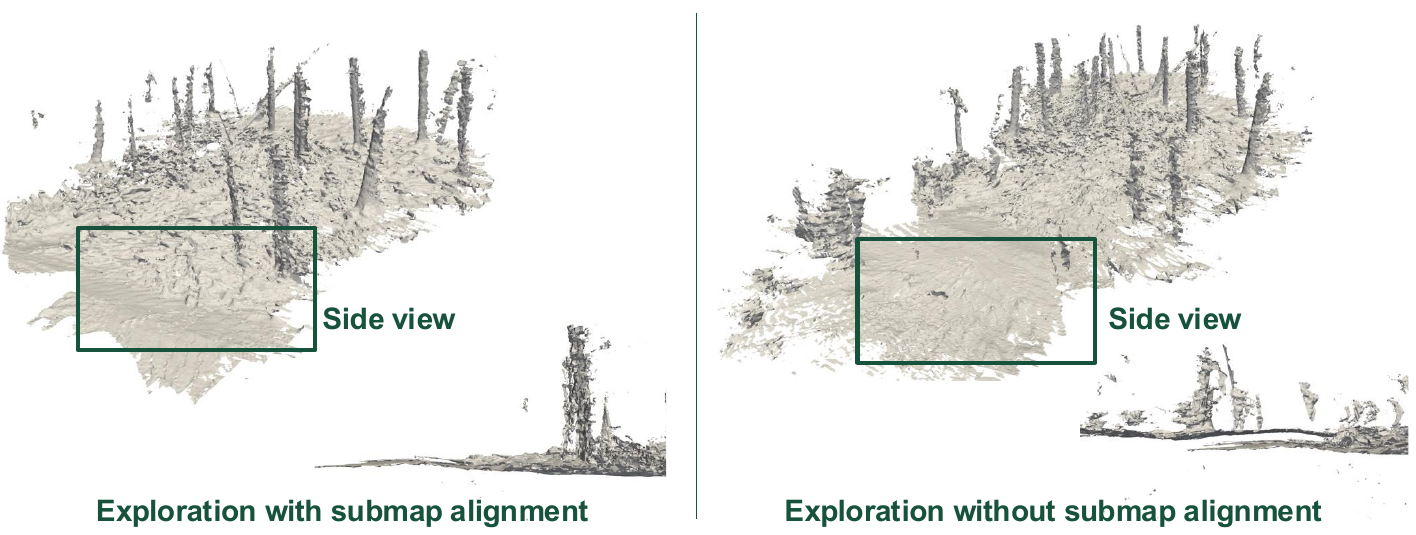}
	\caption{3D mesh reconstructed onboard by MRL drone in the field trip, Stein am Rhein. (Left) The first and the last submaps are well-aligned due to submap alignment factors in the factor graph optimization. (Right) The submaps are geometrically inconsistent without the submap alignment factor.}
	\label{fig:srldrone_forest}
\end{figure*}

To demonstrate the benefits of incorporating geometric information into the state estimation, we performed two missions in the same area, one leveraging the submap constraints and another without them. In both missions, the depth was obtained from the RealSense D455. The mission was an exploration setup, and after 10 minutes (the UAV's battery lifetime), the UAV was manually flown to the start position. Since the end and start area are the same, we perform an ICP alignment to these submaps and estimate the error by the translation of the final transform. Even though the robots start and finish in the same area, there were no visual overlaps since the agents had opposing viewing directions, impeding loop-closures. The approach that had enabled the submap constraints achieved an end position error of \SI{0.19}{\meter} while without submap constraints the final error was \SI{0.37}{\meter}. A qualitative example of the effect of the submap constraints in the mesh reconstruction can be seen in \figref{fig:srldrone_forest}. At the same time, we demonstrate that the proposed approach can effectively work with both depth from LiDAR or from cameras.

When comparing both approaches, \figref{fig:blk_reconstruction} and \figref{fig:srldrone_forest}, it is clear that the LiDAR approach produces a more accurate and complete reconstruction thanks to its larger field-of-view and the unmatched depth-sensing capabilities of this sensor. On average, the missions that relied on visual perception mapped an estimated volume of \SI{706.69}{\cubic\meter}, while with the BLK2FLY mapped volume is of \SI{8612.32}{\cubic\meter}.

A total of 10 autonomous exploration missions were performed, where no manual intervention by the safety pilot was required. This demonstrates that the proposed system is safe and can be used in forest scenarios, even when densely cluttered.

\subsection{Panoptic Segmentation}
For quantitative experimental evaluation, we used our dataset presented in Sec.~\ref{sec:ubonn-dataset}.
The dataset is split into a training dataset comprising 2 complete plots over all time points and the remaining plots from the first measurement campaign. For validation, i.e., hyper-parameter tuning, we use a complete temporal sequence of a single plot.
For evaluation of the performance on a separate test set, we use a plot containing a mixture of tree species and distributions to test the generalization performance of trained models.
The clear spatial separation of train, validation, and test split allows for benchmarking methods in an unbiased fashion with reduced risk of overfitting.

\begin{table*}[t]
  \centering
  \caption{Results for semantic interpretation on the test set of DigiForests. PQ is computed considering the ground, shrub, and tree classes.}
  \begin{tabular}{P{3.5cm}C{1.2cm}C{1.2cm}C{1.2cm}C{1.2cm}C{1.2cm}C{1.2cm}}
    \toprule
    Approach &  Ground & Shrub & Stem & Canopy & Tree & PQ   \\
    \midrule
    MaskPLS \cite{marcuzzi2023ral} & 65.9 & 53.9 & - & - & 57.0 & 58.9 \\
    \textbf{Our approach}  & \textbf{79.5} & \textbf{72.7} & \textbf{78.5} & \textbf{48.2} & \textbf{57.8}  & \textbf{70.0} \\
    \bottomrule
  \end{tabular}
  \label{tab:segresults}
\end{table*}

For evaluating panoptic segmentation performance, we use the panoptic quality metric~\cite{kirillov2019cvpr-ps, porzi2019cvpr}, where the per-class panoptic quality~$\text{PQ}_c$ is given by
\begin{align}\label{eq:pq_c}
  \text{PQ}_c & = \left\{
    \begin{array}{cl}
      \dfrac{
        \sum_{(p, g) \in \text{TP}} \text{IoU}(p, g)
      }{
        |\text{TP}| + \frac{1}{2}|\text{FP}|  + \frac{1}{2}|\text{FN}|
      } &, \text{if $c$ is thing} \\ [0.5cm]
      \text{IoU}(p,g) &, \text{if $c$ is stuff}
    \end{array}
  \right.,
\end{align}
where $p$ is the predicted and $g$ is the ground truth segment of a class $c$, TP is the set of true positives, FP is the set of false positives, FN is the set of false negatives.  With $|\cdot|$, we refer to the number of elements in the given set.
For so-called thing classes, we follow the data assignment by Kirillov et al.~\cite{kirillov2019cvpr-ps}, which treats each predicted instance segment $p$ as a true positive if a ground truth segment $g$ has an intersection over union (IoU) greater than $0.5$.
Conversely,  predicted instance segments $p$ without a matching ground truth segment are false positives, and ground truth segments $g$ without an associated prediction are false negatives.
For stuff classes, we compute the IoU between all points $p$ assigned to class $c$ and the points $g$ with class $c$ in the ground truth.
The overall panoptic quality PQ over all classes $\mathcal{C}$, \ie, ground, tree, and shrub, are given by the average over class-wise panoptic qualities $\text{PQ}_c$ given by:
\begin{align}\label{eq:avgpq}
  \text{PQ} & = \frac{1}{|\mathcal{C}|}\sum_{c\in \mathcal{C}} \text{PQ}_c \, .
\end{align}

\begin{figure}[t]
	\centering
  \includegraphics[width=\linewidth]{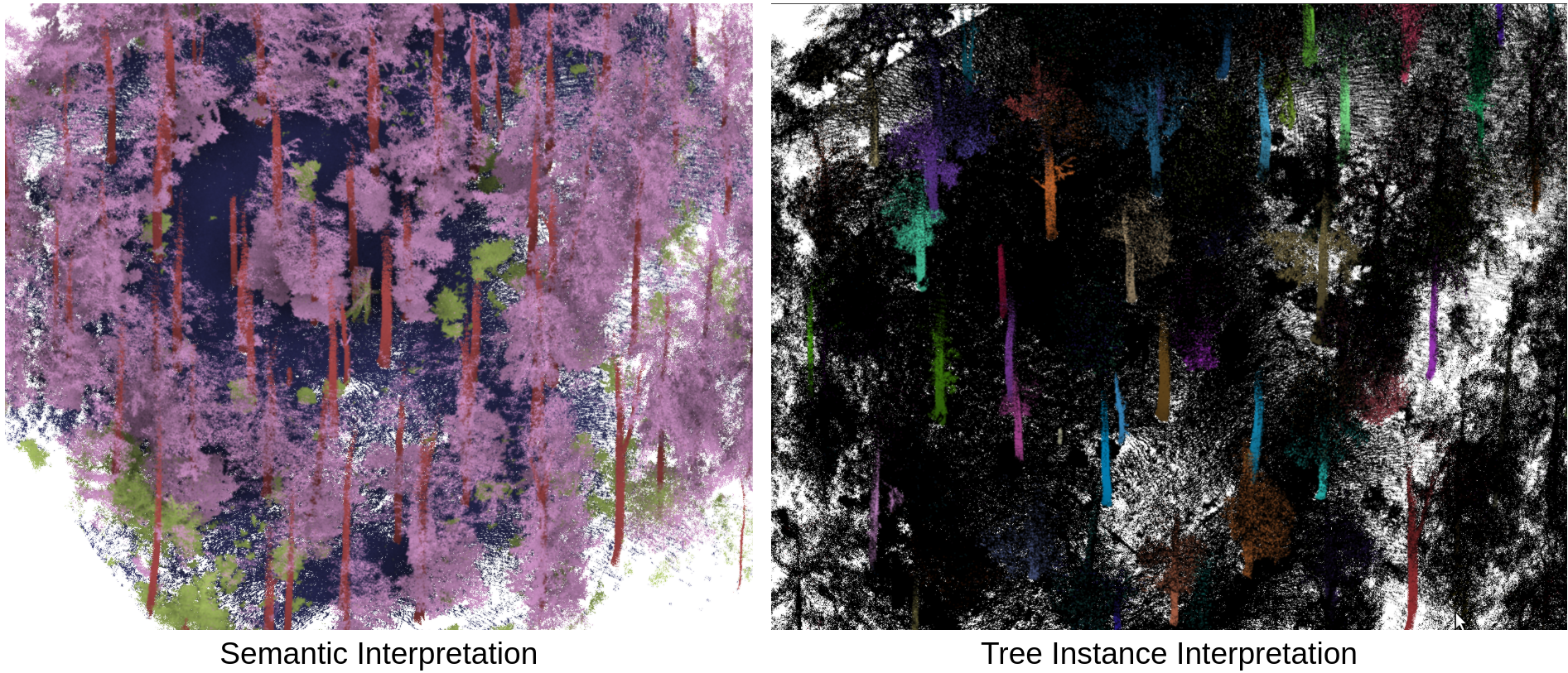}
	\caption{Exemplary semantic and instance predictions generated with our panoptic segmentation approach showing the semantic classes: ground (blue), shrub (green), \textcolor{treecolor}canopy (purple), and stem (ground). In the right part, the instance predictions are shown via a random coloring, where tree instances get the same color assigned.}
	\label{fig:qualitative_panoptic}
\end{figure}

Tab.~\ref{tab:segresults} presents the test set results of our approach compared to a LiDAR-based panoptic segmentation approach, called MaskPLS~\cite{marcuzzi2023ral}, which was developed in the context of autonomous driving using a mask-based Transformer decoder to generate per-class binary masks.
In comparison, our approach can also predict simultaneously the stem and canopy classes, which overlap with the tree class as explained in Sec.~\ref{sec:ubonn-dataset}. Furthermore, our approach shows promising performance in comparison to the Transformer-based architecture, which might be caused by the amount of available training data.
A qualitative example of the predictions is shown in Fig.~\ref{fig:qualitative_panoptic}.

The distinction of stem and canopy by the offline processing enables tackling scenarios where pure geometric approaches struggle due to small bushes and other vegetation close to the trees.
The more fine-grained segmentation enables a more complete forest inventory, and enables deriving other forestry-related traits, such as the clear wood percentage of the tree by using the stem region, the canopy volume due to the separation of the foliage, or the more complete estimation of the diameter at breast height.
Tab.~\ref{tab:dbh_results} shows the advantage with respect to the geometric online estimation of DBH onboard a robotic system~\cite{Freissmuth2024} described in Sec.~\ref{sec:forest-inventory-legged}
In particular, the online approach cannot reliably handle some cases, where the offline processing and a more fine-grained distinction into stem segments can provide a DBH estimate, as shown by the higher per-plot average recall and overall recall.

\begin{table}[t]
  \setlength\tabcolsep{5pt}
  \centering
  \caption{Results for tree DBH estimation using the DigiForests dataset. Root mean squared error (RMSE) is reported in cm.}
  \begin{tabular}{P{3.3cm}C{1.6cm}C{1.6cm}C{1.6cm}C{1.6cm}}
    \toprule
    & \multicolumn{2}{c}{Recall} & \multicolumn{2}{c}{RMSE} \\
    \cmidrule(lr){2-3}
    \cmidrule(lr){4-5}
    Approach &Per-plot Avg. & Overall & Per-plot Avg. & Overall \\
    \midrule
    Geometric approach~\cite{Freissmuth2024} & 0.88 & 0.92 & \textbf{3.42} & \textbf{3.15} \\
    Cylinder fitting on\newline stem segments & \textbf{0.98} & \textbf{0.98} & 5.32 & 6.41 \\
    \bottomrule
  \end{tabular}
  \label{tab:dbh_results}
  \vspace{-3mm}  %
\end{table}

In summary, %
our approach shows promising results compared to conventional LiDAR-based panoptic segmentation approaches employed in the context of autonomous driving. The approach is tailored for offline processing of aggregated scans, and a promising avenue for future work is an extension towards single scan panoptic segmentation and integration into a mapping framework to allow building semantic maps on board a robotic system.

\section{Conclusions and Future Work}
\label{sec:conclusion}
In this chapter, we have presented DigiForest, an innovative approach towards large-scale precision forestry that addresses the whole process of forest management, from data collection by a team of heterogeneous robots, its processing and transformation into a forest inventory, the integration of the inventory into a decision support system, and finally the autonomous tree harvesting via a driverless, purpose-build, hydraulic machine.

All these technologies have been validated in real operative scenarios, such as the Oberwald forest in Switzerland and other locations in Finland and the UK.  The field trials served to test and validate key functionalities under realistic conditions, highlighting both the potential and the remaining technical challenges. Rather than replacing traditional forestry practices, the aim is to support them with more reliable, adaptable, and data-informed tools.

These developments reflect steady progress toward the integration of robotic perception and autonomy in forestry operations. These systems are being designed with forest-specific challenges in mind, including variable terrain, dense vegetation, and limited visibility.

Future work will develop across two main directions: making each individual component more robust (e.g., by testing in more challenging scenarios) and increasing the integration between components. For example, we plan to enhance the SAHA harvester's global route planning and reactive obstacle avoidance by using real‐time semantic terrain analysis to classify ground cover and hazards.

\printbibliography

@string{ijrr  = "Int. J. Robot. Res."}

@string{ral   = "{IEEE} Robot. Autom. Lett."}

@string{scirob = "Sci. Robot."}

@string{tro   = "{IEEE} Trans. Robot."}

@string{ieee  = "Proc. of the IEEE"}

@string{pami  = "{IEEE} Trans. Pattern Anal. Mach. Intell."}

@string{cvpr  = "{IEEE} Int. Conf. Computer Vision and Pattern Recognition (CVPR)"}

@string{iccv  = "Intl. Conf. on Computer Vision (ICCV)"}

@string{icra  = "IEEE Int. Conf. Robot. Autom. (ICRA)"}

@string{iros  = "{IEEE/RSJ} Int. Conf. Intell. Robots Syst. (IROS)"}

@article{Mattamala2022,
  author  = {Mat{\'{\i}}as Mattamala and
             Nived Chebrolu and
             Maurice F. Fallon},
  doi     = {10.1109/LRA.2022.3143196},
  journal = ral,
  number  = {2},
  pages   = {2353--2360},
  title   = {An Efficient Locally Reactive Controller for Safe Navigation in Visual
             Teach and Repeat Missions},
  volume  = {7},
  year    = {2022}
}

@inproceedings{Freissmuth2024,
  author    = {Leonard Frei{\ss}muth and Matias Mattamala and Nived Chebrolu and Simon Schaefer and Stefan Leutenegger and Maurice Fallon},
  booktitle = iros,
  title     = {{Online Tree Reconstruction and Forest Inventory on a Mobile~Robotic~System}},
  year      = {2024}
}

@article{Wisth2023,
  author  = {Wisth, David and Camurri, Marco and Fallon, Maurice},
  doi     = {10.1109/TRO.2022.3193788},
  journal = tro,
  number  = {1},
  pages   = {309-326},
  title   = {VILENS: Visual, Inertial, Lidar, and Leg Odometry for All-Terrain Legged Robots},
  volume  = {39},
  year    = {2023}
}

@article{Besl1992,
  author  = {Paul J. Besl and
             Neil D. McKay},
  doi     = {10.1109/34.121791},
  journal = {{IEEE} Trans. Pattern Anal. Mach. Intell.},
  number  = {2},
  pages   = {239--256},
  title   = {A Method for Registration of 3-D Shapes},
  volume  = {14},
  year    = {1992}
}

@inproceedings{Ramezani2020a,
  author    = {Ramezani, Milad and Tinchev, Georgi and Iuganov, Egor and Fallon,
               Maurice},
  booktitle = icra,
  doi       = {10.1109/ICRA40945.2020.9196769},
  number    = {},
  pages     = {4158-4164},
  title     = {{Online LiDAR-SLAM for Legged Robots with Robust Registration and Deep-Learned Loop Closure}},
  volume    = {},
  year      = {2020}
}

@article{Kaess2012,
  author  = {Michael Kaess and
             Hordur Johannsson and
             Richard Roberts and
             Viorela Ila and
             John J. Leonard and
             Frank Dellaert},
  doi     = {10.1177/0278364911430419},
  journal = ijrr,
  number  = {2},
  pages   = {216--235},
  title   = {{iSAM2: Incremental smoothing and mapping using the Bayes tree}},
  volume  = {31},
  year    = {2012}
}

@article{Forster2017,
  author  = {Forster, Christian and Carlone, Luca and Dellaert, Frank and Scaramuzza, Davide},
  doi     = {10.1109/TRO.2016.2597321},
  journal = tro,
  number  = {1},
  pages   = {1-21},
  title   = {On-Manifold Preintegration for Real-Time Visual--Inertial Odometry},
  volume  = {33},
  year    = {2017}
}

@article{Miki2022a,
  author  = {Takahiro Miki  and Joonho Lee  and Jemin Hwangbo  and Lorenz Wellhausen  and Vladlen Koltun  and Marco Hutter },
  doi     = {10.1126/scirobotics.abk2822},
  journal = scirob,
  number  = {62},
  title   = {Learning Robust Perceptive Locomotion for Quadrupedal Robots in the Wild},
  volume  = {7},
  year    = {2022}
}

@article{Duncanson2020,
  title={Global aboveground biomass product validation best practices protocol. Version 1.0},
  author={Duncanson, L and Armston, J and Disney, M and Avitabile, V and Barbier, N and Calders, K and Carter, S and Chave, J and Herold, M and Macbean, N and others},
  journal={Best practice protocol for satellite derived land product validation. Land Product Validation Subgroup (Working Group on Calibration and Validation, Committee on Earth Observation Satellites)},
  year={2020}
}

@inproceedings{Miki2022b,
  author    = {Miki, Takahiro and
               Wellhausen, Lorenz and
               Grandia, Ruben and
               Jenelten, Fabian and
               Homberger, Timon and
               Hutter, Marco},
  booktitle = iros,
  doi       = {10.1109/IROS47612.2022.9981507},
  pages     = {2273-2280},
  title     = {Elevation Mapping for Locomotion and Navigation using GPU},
  year      = {2022}
}

@inproceedings{behley2019iccv,
author = {J. Behley and M. Garbade and A. Milioto and J. Quenzel and S. Behnke and C. Stachniss and J. Gall},
title = {{SemanticKITTI: A Dataset for Semantic Scene Understanding of LiDAR Sequences}},
booktitle = iccv,
year = {2019},
}

@inproceedings{
jelavic2022harveri,
title={Harveri : A Small (Semi-)Autonomous Precision Tree Harvester},
author={Edo Jelavic and Tun Kapgen and Simon Kerscher and Dominic Jud and Marco 
Hutter},
booktitle={ICRA 2022 Workshop in Innovation in Forestry Robotics: Research and 
Industry Adoption},
year={2022},
url={https://openreview.net/forum?id=BOdgBn6OQM5}
}

@inproceedings{malladi2025icra,
  author    = {Meher V.R. Malladi and Nived Chebrolu and Irene Scacchetti and Luca Lobefaro and Tiziano Guadagnino and Beno\^{i}t Casseau
    and Haedam Oh and Leonard Frei{\ss}muth and Markus Karppinen and Janine Schweier and Stefan Leutenegger and Jens Behley
    and Cyrill Stachniss and Maurice Fallon},
  title     = {{DigiForests: A Longitudinal LiDAR Dataset for Forestry Robotics}},
  booktitle   = icra,
  year      = {2025},
}

@article{leutenegger2022okvis2,
	title={Okvis2: Realtime scalable visual-inertial slam with loop closure},
	author={Leutenegger, Stefan},
	journal={arXiv preprint arXiv:2202.09199},
	year={2022}
}

@INPROCEEDINGS{bocheLVIOkvis,
	author={Boche, Simon and Laina, Sebastián Barbas and Leutenegger, Stefan},
	booktitle=icra, 
	title={Tightly-Coupled LiDAR-Visual-Inertial SLAM and Large-Scale Volumetric Occupancy Mapping}, 
	year={2024},
	pages={18027-18033},
	doi={10.1109/ICRA57147.2024.10610460}}

@article{funk2021multi,
	title={Multi-resolution 3D mapping with explicit free space representation for fast and accurate mobile robot motion planning},
	author={Funk, Nils and Tarrio, Juan and Papatheodorou, Sotiris and Popovi{\'c}, Marija and Alcantarilla, Pablo F and Leutenegger, Stefan},
	journal=ral,
	volume={6},
	number={2},
	pages={3553--3560},
	year={2021},
}

@article{xu2023unifying,
  title={Unifying flow, stereo and depth estimation},
  author={Xu, Haofei and Zhang, Jing and Cai, Jianfei and Rezatofighi, Hamid and Yu, Fisher and Tao, Dacheng and Geiger, Andreas},
  journal=pami,
  volume={45},
  number={11},
  pages={13941--13958},
  year={2023},
}

@inproceedings{gammell2014informed,
  title={Informed RRT*: Optimal sampling-based path planning focused via direct sampling of an admissible ellipsoidal heuristic},
  author={Gammell, Jonathan D and Srinivasa, Siddhartha S and Barfoot, Timothy D},
  booktitle=iros,
  pages={2997--3004},
  year={2014},
}

@article{papatheodorou2024efficient,
  title={Efficient Submap-based Autonomous MAV Exploration using Visual-Inertial SLAM Configurable for LiDARs or Depth Cameras},
  author={Papatheodorou, Sotiris and Boche, Simon and Laina, Sebasti{\'a}n Barbas and Leutenegger, Stefan},
  journal={arXiv preprint arXiv:2409.16972},
  year={2024}
}

@article{hutter2016force,
  title={Force control for active chassis balancing},
  author={Hutter, Marco and Leemann, Philipp and Hottiger, Gabriel and Figi, Ruedi and Tagmann, Stefan and Rey, Gonzalo and Small, George},
  journal={IEEE/ASME Transactions On Mechatronics},
  volume={22},
  number={2},
  pages={613--622},
  year={2016},
  publisher={IEEE}
}

@article{nan2024learning,
  title={Learning adaptive controller for hydraulic machinery automation},
  author={Nan, Fang and Hutter, Marco},
  journal=ral,
  year={2024},
}

@inproceedings{hu2024motion,
  title={Motion Primitives Planning For Center-Articulated Vehicles},
  author={Hu, Jiangpeng and Yang, Fan and Nan, Fang and Hutter, Marco},
  booktitle=iros,
  pages={12702--12709},
  year={2024},
}

@inproceedings{ester1996kdd,
  title = {A density-based algorithm for discovering clusters in large spatial
           databases with noise.},
  author = {M. Ester and H. Kriegel and J. Sander and X.Xu},
  booktitle = {Proc. of the Conf. on Knowledge Discovery and Data Mining (KDD)},
  year = {1996},
}

@inproceedings{choy2019cvpr,
  author = {C. Choy and J. Gwak and S. Savarese},
  title = {{4D Spatio-Temporal ConvNets: Minkowski Convolutional Neural Networks}},
  booktitle = cvpr,
  year = 2019,
}

@INPROCEEDINGS{casseau2024iros,
  author={Casseau, Benoît and Chebrolu, Nived and Mattamala, Matias and Freissmuth, Leonard and Fallon, Maurice},
  booktitle=iros, 
  title={{Markerless Aerial-Terrestrial Co-Registration of Forest Point Clouds using a Deformable Pose Graph}}, 
  year={2024},
  pages={39-46},
  doi={10.1109/IROS58592.2024.10802448}}

@mastersthesis{depetris2020msc,
    author = "Paolo De Petris",
    title = "The Resilient Micro Flyer: a New Collision-tolerant Autonomous Robot",
    school = "University of Nevada, Reno",
    year = "2020"
}

@article{Bugmann,
author = {Bugmann, Harald K. M.},
title = {A Simplified Forest Model to Study Species Composition Along Climate Gradients},
journal = {Ecology},
volume = {77},
number = {7},
pages = {2055-2074},
doi = {https://doi.org/10.2307/2265700},
url = {https://esajournals.onlinelibrary.wiley.com/doi/abs/10.2307/2265700},
eprint = {https://esajournals.onlinelibrary.wiley.com/doi/pdf/10.2307/2265700},
year = {1996}
}

@misc{EU2021,
  author    = {European Commission},
  publisher = {European Commision Brussels},
  title     = {{Commission communication: New EU forest strategy for 2030}},
  url       = {https://eur-lex.europa.eu/legal-content/EN/TXT/?uri=CELEX%3A52021DC0572},
  year      = {2021}
}

@ARTICLE{Corke2001,
  author={Corke, P.I. and Ridley, P.},
  journal={IEEE Transactions on Robotics and Automation}, 
  title={Steering kinematics for a center-articulated mobile robot}, 
  year={2001},
  volume={17},
  number={2},
  pages={215-218},
  doi={10.1109/70.928568}}

@inproceedings{ronneberger2015micc,
  author    = {O. Ronneberger and P.Fischer and T. Brox},
  title     = {{U-Net: Convolutional Networks for Biomedical Image Segmentation}},
  booktitle = {Proc. of the Medical Image Computing and Computer-Assisted Intervention (MICCAI)},
  year      = 2015,
  url       = {https://arxiv.org/pdf/1505.04597.pdf}
}

@article{mattamala2025tfr,
  title        = {Building Forest Inventories With Autonomous Legged Robots—System, Lessons, and Challenges Ahead},
  author       = {Mattamala, Matías and Chebrolu, Nived and Frey, Jonas and Freißmuth, Leonard and Oh, Haedam and Casseau, Benoit and Hutter, Marco and Fallon, Maurice},
  year         = 2025,
  journal      = {IEEE Transactions on Field Robotics},
  volume       = 2,
  number       = {},
  pages        = {418--436},
  doi          = {10.1109/TFR.2025.3583972}
}

@inproceedings{porzi2019cvpr,
  author    = {L. Porzi and S. Rota Bulo and A. Colovic and P. Kontschieder},
  title     = {{Seamless Scene Segmentation}},
  booktitle = cvpr,
  year      = 2019,
}

@inproceedings{kirillov2019cvpr-ps,
  author    = {A. Kirillov and K. He and R. Girshick and C. Rother and P. Doll{\'a}r},
  title     = {{Panoptic Segmentation}},
  booktitle = cvpr,
  year      = 2019,
}

@article{marcuzzi2023ral,
  author  = {R. Marcuzzi and L. Nunes and L. Wiesmann and J. Behley and C. Stachniss},
  title   = {{Mask-Based Panoptic LiDAR Segmentation for Autonomous Driving}},
  journal = ral,
  volume  = {8},
  number  = {2},
  pages   = {1141--1148},
  year    = 2023,
}

@INPROCEEDINGS{Frey2022Traversability,
  author={Frey, Jonas and Hoeller, David and Khattak, Shehryar and Hutter, Marco},
  booktitle={2022 IEEE/RSJ International Conference on Intelligent Robots and Systems (IROS)}, 
  title={Locomotion Policy Guided Traversability Learning using Volumetric Representations of Complex Environments}, 
  year={2022},
  volume={},
  number={},
  pages={5722-5729},
  doi={10.1109/IROS47612.2022.9982190}}

@inproceedings{oh2024iros,
  title        = {Evaluation and Deployment of LiDAR-based Place Recognition in Dense Forests},
  author       = {Oh, Haedam and Chebrolu, Nived and Mattamala, Matias and Freißmuth, Leonard and Fallon, Maurice},
  year         = 2024,
  booktitle    = iros,
  volume       = {},
  number       = {},
  pages        = {12824--12831},
  doi          = {10.1109/IROS58592.2024.10801297}
}

@inproceedings{vidanapathirana2022icra,
  author={Vidanapathirana, Kavisha and Ramezani, Milad and Moghadam, Peyman and Sridharan, Sridha and Fookes, Clinton},
  booktitle=icra, 
  title={LoGG3D-Net: Locally Guided Global Descriptor Learning for 3D Place Recognition}, 
  year={2022},
  volume={},
  number={},
  pages={2215-2221},
  doi={10.1109/ICRA46639.2022.9811753}}

@inproceedings{fu2023iros,
  author={Fu, Lanke Frank Tarimo and Chebrolu, Nived and Fallon, Maurice},
  booktitle=iros, 
  title={Extrinsic Calibration of Camera to LIDAR Using a Differentiable Checkerboard Model}, 
  year={2023},
  volume={},
  number={},
  pages={1825-1831},
  doi={10.1109/IROS55552.2023.10341781}}

@INPROCEEDINGS{furgale2013iros,
  author={Furgale, Paul and Rehder, Joern and Siegwart, Roland},
  booktitle=iros, 
  title={Unified temporal and spatial calibration for multi-sensor systems}, 
  year={2013},
  volume={},
  number={},
  pages={1280-1286},
  doi={10.1109/IROS.2013.6696514}}

@article{patacca2023gcb,
  title        = {Significant increase in natural disturbance impacts on European forests since 1950},
  author       = {Patacca, Marco and Lindner, Marcus and Lucas-Borja, Manuel Esteban and Cordonnier, Thomas and Fidej, Gal and Gardiner, Barry and Hauf, Ylva and Jasinevičius, Gediminas and Labonne, Sophie and Linkevičius, Edgaras and Mahnken, Mats and Milanovic, Slobodan and Nabuurs, Gert-Jan and Nagel, Thomas A. and Nikinmaa, Laura and Panyatov, Momchil and Bercak, Roman and Seidl, Rupert and Ostrogović Sever, Masa Zorana and Socha, Jaroslaw and Thom, Dominik and Vuletic, Dijana and Zudin, Sergey and Schelhaas, Mart-Jan},
  year         = 2023,
  journal      = {Global Change Biology},
  volume       = 29,
  number       = 5,
  pages        = {1359--1376},
  doi          = {https://doi.org/10.1111/gcb.16531}
}

@article{treeseg,
  title        = {Extracting individual trees from lidar point clouds using treeseg},
  author       = {Burt, Andrew and Disney, Mathias and Calders, Kim},
  year         = 2019,
  journal      = {Methods in Ecology and Evolution},
  volume       = 10,
  number       = 3,
  pages        = {438--445},
  doi          = {https://doi.org/10.1111/2041-210X.13121}
}

@article{bauewns2016forests,
  title        = {Forest Inventory with Terrestrial LiDAR: A Comparison of Static and Hand-Held Mobile Laser Scanning},
  author       = {Bauwens, Sébastien and Bartholomeus, Harm and Calders, Kim and Lejeune, Philippe},
  year         = 2016,
  journal      = {Forests},
  volume       = 7,
  number       = 6,
  doi          = {10.3390/f7060127},
  issn         = {1999-4907},
  note         = 127
}

@misc {foresthandbook,
title = "Forest Mensuration Handbook",
author= "G. J. Hamilton",
year = 1988
}

@inproceedings{slam2007harvester,
  title        = {Simultaneous Localization and Mapping for Forest Harvesters},
  author       = {Miettinen, Mikko and Ohman, Matti and Visala, Arto and Forsman, Pekka},
  year         = 2007,
  booktitle    = {Proceedings 2007 IEEE International Conference on Robotics and Automation},
  pages        = {517--522},
  doi          = {10.1109/ROBOT.2007.363838}
}

@article{lidarremotesensing,
  title        = {Lidar Remote Sensing for Forestry},
  author       = {Dubayah, Ralph O. and Drake, Jason B.},
  year         = 2000,
  month        = {06},
  journal      = {Journal of Forestry},
  volume       = 98,
  number       = 6,
  pages        = {44--46},
  doi          = {10.1093/jof/98.6.44},
  issn         = {0022-1201}
}

@article{pierzchala2018compag,
  title        = {Mapping forests using an unmanned ground vehicle with 3D LiDAR and graph-SLAM},
  author       = {Marek Pierzchała and Philippe Giguère and Rasmus Astrup},
  year         = 2018,
  journal      = {Computers and Electronics in Agriculture},
  volume       = 145,
  pages        = {217--225},
  doi          = {https://doi.org/10.1016/j.compag.2017.12.034},
  issn         = {0168-1699}
}

@article{pomerlau2020jfr,
  title        = {Automatic three-dimensional mapping for tree diameter measurements in inventory operations},
  author       = {Tremblay, Jean-François and Béland, Martin and Gagnon, Richard and Pomerleau, François and Giguère, Philippe},
  year         = 2020,
  journal      = {Journal of Field Robotics},
  volume       = 37,
  number       = 8,
  pages        = {1328--1346},
  doi          = {https://doi.org/10.1002/rob.21980}
}

@techreport{payloadapi,
  title        = {Report defining data format and software API for multi-robot map sharing},
  author       = {Nived Chebrolu and Maurice Fallon},
  year         = 2023,
  institution  = {University of Oxford}
}

@misc{zacharia2025ral,
      title={Collaborative Exploration with a Marsupial Ground-Aerial Robot Team through Task-Driven Map Compression}, 
      author={Angelos Zacharia and Mihir Dharmadhikari and Kostas Alexis},
      year={2025},
      eprint={2509.07655},
      archivePrefix={arXiv},
      primaryClass={cs.RO},
      url={https://arxiv.org/abs/2509.07655}, 
}

\end{document}